%% file: 2025_arr_audio.tex
\pdfoutput=1

\documentclass[11pt]{article}

\newif\ifcomment\commenttrue
\input{style/preamble}

\usepackage[final]{style/acl}
\usepackage{times}
\usepackage{latexsym}
\usepackage{booktabs}
\usepackage{tabularx}
\usepackage{array}

\usepackage[T1]{fontenc}

\usepackage[utf8]{inputenc}

\usepackage{microtype}

\usepackage{inconsolata}

\usepackage{graphicx}
\usepackage{siunitx}

\newcommand{\name}[0]{\abr{Audita}}
\newcommand{\acronym}[0]{{\textbf{A}udio \textbf{U}nderstanding from \textbf{D}iverse \textbf{I}nternet \textbf{T}rivia \textbf{A}uthors}}
\newcommand{\HUMANACCURACY}[0]{\abr{$32.13$}}
\newcommand{\HUMANMCQACCURACY}[0]{\abr{$60.16$}}
\newcommand{\MODELACCURACY}[0]{\abr{$8.86$}}
\newcommand{\q}[1]{\textit{``#1''}}
\newcommand{\ans}[1]{\underline{\textit{#1}}}

\newcommand{\MODELIRTSCORE}[0]{\abr{$-2.79$}}
\newcommand{\HUMANIRTSCORE}[0]{\abr{$0.05$}}

\newcommand{\aqa}[0]{\mbox{Audio QA}}
\title{\name{}: A New Dataset to Audit Humans vs.~AI Skill at Audio QA}

\author{Tasnim Kabir \\
University of Maryland \\
  \texttt{tkabir1@umd.edu} 
  \\\And
    Dmytro Kurdydyk 
    \\Davidson College\\\texttt{dmkurdydyk@davidson.edu}\\\And Aadi Palnitkar\\ 
University of Maryland\\ \texttt{apalnitk@terpmail.umd.edu}\\\AND Liam Dorn 
\\Columbia University\\\texttt{lmd2243@columbia.edu}\\\And Ahmed Haj Ahmed \\ 
Haverford College\\\texttt{ahajahmed@haverford.edu}\\\And
  Jordan Lee Boyd-Graber \\
 University of Maryland \\
  \texttt{jbg@umiacs.umd.edu}}

\begin{document}
\maketitle
\input{2025_arr_audio/sections/00-abstract.tex}
\input{2025_arr_audio/sections/10-introduction}
\input{2025_arr_audio/sections/20-datasets_matter}
\input{2025_arr_audio/sections/30-dataset}
\input{2025_arr_audio/sections/40-experiments}
\input{2025_arr_audio/sections/50-analysis}
\input{2025_arr_audio/sections/51-error}
\input{2025_arr_audio/sections/55-related_work}
\input{2025_arr_audio/sections/60-conclusion}
\input{2025_arr_audio/sections/70-acknowledgements}

\bibliography{bib/custom}

\appendix
\input{2025_arr_audio/sections/appendix-related_work}



\end{document}

%% file: style/preamble.tex
\usepackage[a-1b]{pdfx}

\usepackage{framed}
\usepackage{mdwlist}
\usepackage{siunitx}
\usepackage{latexsym}
\usepackage{colortbl}
\usepackage{xcolor}
\usepackage{booktabs}
\usepackage{fnpct}
\usepackage{amsfonts}
\usepackage[T1]{fontenc}
\usepackage{bold-extra}
\usepackage{amsmath}
\usepackage{amssymb}
\usepackage{bm}
\usepackage{graphicx}
\usepackage{mathtools}
\usepackage{microtype}
\usepackage{multirow}
\usepackage{multicol}
\usepackage{xpatch}
\usepackage{latexsym,comment}
\usepackage[normalem]{ulem}

\newcommand*{\missingreference}{{\Huge \colorbox{red}{?reference?}}}
\newcommand*{\missingcitation}{{\Huge \colorbox{red}{?citation?}}}

\makeatletter
\xpatchcmd{\@setref}{\bfseries}{\missingreference}{}{}
\def\@citex[#1]#2{\leavevmode
    \let\@citea\@empty
    \@cite{\@for\@citeb:=#2\do
        {\@citea\def\@citea{,\penalty\@m\ }%
            \edef\@citeb{\expandafter\@firstofone\@citeb\@empty}%
            \if@filesw\immediate\write\@auxout{\string\citation{\@citeb}}\fi
            \@ifundefined{b@\@citeb}{\hbox{\reset@font\missingcitation}%
                \G@refundefinedtrue
                \@latex@warning
                {Citation `\@citeb' on page \thepage \space undefined}}%
            {\@cite@ofmt{\csname b@\@citeb\endcsname}}}}{#1}}
\makeatother

\newcommand{\gem}[1]{\mbox{\textsc{gem}}}
\newcommand{\abr}[1]{\textsc{#1}}

\renewenvironment{quote}
{\list{}{\rightmargin\leftmargin}%
    \item\relax\small\ignorespaces}
{\unskip\unskip\endlist}

\newcommand{\hidetext}[1]{}
\newcommand{\ignore}[1]{}

\ifcomment
    \newcommand{\pinaforecomment}[3]{\colorbox{#1}{\parbox{.8\linewidth}{#2: #3}}}

    \newcommand{\prtodo}[1]{\pinaforecomment{lightblue}{pr}{#1}}
    \newcommand{\prtodoi}[1]{\pinaforecomment{lightblue}{pr}{#1}}
\else
    \newcommand{\pinaforecomment}[3]{}
    \newcommand{\prtodo}[1]{}
    \newcommand{\prtodoi}[1]{}
\fi

\newcommand{\jbgcomment}[1]{\pinaforecomment{red}{JBG}{#1}}

\newcommand{\smallurl}[1]{ \begin{tiny}\url{#1}\end{tiny}}

\definecolor{lightblue}{HTML}{3cc7ea}
\definecolor{CUgold}{HTML}{CFB87C}
\definecolor{grey}{rgb}{0.95,0.95,0.95}
\definecolor{ceil}{rgb}{0.57, 0.63, 0.81}
\definecolor{UMDred}{HTML}{ed1c24}
\definecolor{UMDyellow}{HTML}{ffc20e}


%% file: 2025_arr_audio/sections/00-abstract.tex
\begin{abstract}
Existing audio question answering benchmarks largely emphasize sound event classification or caption-grounded queries, often enabling models to succeed through shortcut strategies, short-duration cues, lexical priors, dataset-specific biases, or even bypassing audio via metadata and captions rather than genuine reasoning 
Thus, we present \name{} (\acronym{}), 
a large-scale, real-world benchmark to rigorously evaluate audio reasoning beyond surface-level acoustic recognition. 
AUDITA\footnote{The codebase and data are available at \url{https://github.com/Pinafore/audio_data}, and~\url{https://huggingface.co/datasets/TasnimKabir12/audita-audio}} comprises carefully curated, human-authored trivia questions grounded in real-world audio, designed to stress robust auditory reasoning through challenging distractors and long-range temporal dependencies, using probing queries that cannot be answered from isolated text or sound cues alone. Human average accuracy of \HUMANACCURACY\% shows both the challenge of the task while demonstrating meaningful comprehension of the audio. In stark contrast, state-of-the-art audio question answering models perform poorly, with average accuracy below \MODELACCURACY\%. Beyond raw accuracy, we apply Item Response Theory (IRT) to estimate latent proficiency, question difficulty, and expose systematic deficiencies of the models and data.
\end{abstract}

%% file: 2025_arr_audio/sections/10-introduction.tex
\section{Introduction}

Question answering (QA) is a central paradigm for evaluating language understanding, with modern benchmarks such as Natural Questions~\cite{kwiatkowski-etal-2019-natural} and large-scale multimodal tasks~\cite{yue2025mmmu} driving rapid progress. Combined with advances in large language models~\cite{chowdhery2023palm,DBLP:journals/corr/abs-2303-08774}, these systems now achieve near- or superhuman performance on many text-based QA tasks.

However, this success does not extend uniformly to audio question answering (\aqa{}), where models must reason over complex and often indirect auditory signals. Despite progress in speech recognition~\cite{DBLP:conf/nips/BaevskiZMA20}, audio tagging~\cite{DBLP:journals/taslp/KongCIWWP20}, and multimodal modeling~\cite{DBLP:conf/icassp/GuzhovRHD22}, current systems still struggle to reason about audio beyond surface-level recognition. 

Existing benchmarks fail to capture this gap. Datasets such as VGGSound~\cite{DBLP:conf/icassp/ChenXVZ20} focus on closed-set classification, while captioning datasets like Clotho~\cite{DBLP:conf/icassp/DrossosLV20} emphasize description rather than reasoning. Many \aqa{} datasets further rely on synthetic scenes or templated questions~\cite{DBLP:journals/taslp/FayekJ20,DBLP:conf/cvpr/LiWTXW022}, enabling models to exploit language priors or shallow cues rather than true auditory grounding (Section~\ref{sec:motivation}).

A key limitation is the absence of systematic human benchmarks, making it difficult to assess whether models are approaching human-level auditory understanding. Recent work has highlighted the need for more reliable evaluation frameworks such as Item Response Theory~\cite[\abr{IRT}]{DBLP:conf/emnlp/LalorWY19}, which jointly models question difficulty and participant ability.

This work introduces \name{} (Section~\ref{sec:dataset}), a large-scale benchmark of human-authored audio trivia questions sourced from real-world domains including quizzes, media, and cultural knowledge. Unlike prior datasets, \name{} avoids synthetic audio and templated formats, focusing instead on crafted questions that require multi-cue auditory reasoning and world knowledge. Under controlled human evaluation (Section~\ref{sec:experiment}), expert participants have strong accuracy (best-per-category:~69.17\% free-form, 86.67\% MCQ), while state-of-the-art audio and multimodal models lag far behind, with accuracies under \MODELACCURACY\%. This indicates a substantial gap in auditory reasoning capability.

We further analyze this gap using IRT~\cite{hambleton1991fundamentals}, revealing a clear separation between human and model ability distributions. Our analysis also surfaces common issues in existing benchmarks (Section~\ref{sec:analysis}): ambiguity, underspecified answers, and shortcut-prone designs, which can inflate model performance without requiring true auditory grounding.

Our contributions are: (1) \textbf{\name{}}, a benchmark for challenging audio QA requiring multi-cue reasoning; (2) \textbf{human baselines} demonstrating a large and consistent human–model gap; and (3) \textbf{IRT-based analysis} providing fine-grained insights into difficulty, reasoning structure, and model failure modes.

%% file: 2025_arr_audio/sections/20-datasets_matter.tex
\section{Why Audio QA Requires Better Questions}\label{sec:motivation}

Recent advances in text-based question answering (QA) and large language model training have highlighted that dataset quality remains a primary bottleneck despite advances in model scale. Poorly designed questions introduce ambiguity~\cite{DBLP:conf/emnlp/MinMHZ20,li2025condambigqa}, false presuppositions~\cite{DBLP:conf/acl/0001MZH23}, unreliable supervision~\cite{DBLP:conf/emnlp/LiMNLB24}, and exploitable shortcuts~\cite{poliak2018hypothesis}, resulting in inflated model performance and unstable evaluations.~\citet{li2025condambigqa,shi-etal-2025-ambiguity} show that a substantial fraction of real-world questions remain inherently ambiguous or underspecified, often leading to apparent hallucinations that stem from dataset limitations rather than model failure. Consequently, contemporary QA research has increasingly emphasized careful question design, ambiguity-aware evaluation, and human calibration to ensure benchmarks truly test reasoning rather than dataset artifacts.

Many of these failure modes identified in text QA also appear in \aqa{} datasets. We analyze these issues through Item Response Theory~\cite[\abr{IRT}]{hambleton1991fundamentals}, a classical framework for modeling question difficulty and discrimination, which captures how effectively questions distinguish between high and low-ability models or annotators (Appendix Section~\ref{IRT}).~\citet{rodriguez2021evaluation} and~\citet{lalor2024item} have applied IRT to benchmark and leaderboard analysis, providing a principled mechanism for diagnosing dataset quality. This diagnostic perspective aligns with more recent efforts toward adaptive and fluid benchmarking that emphasize maintaining meaningful evaluation as models improve~\cite{hofmann2025fluid}. Appendix Table~\ref{tab:dataset_issues_real} and~\ref{tab:dataset_issues_real_improved2} summarize common issues across several popular \aqa{} datasets, illustrating these problems with concrete examples and IRT-based evidence. Ambiguity, a central challenge extensively studied in \cite{DBLP:conf/emnlp/MinMHZ20}, occurs when questions admit multiple plausible answers or interpretations. For example, in Clotho-\aqa{}~\cite{DBLP:conf/icassp/DrossosLV20}, \q{What animal is making the sound?} receives different yet valid responses such as \ans{bird} or \ans{dog}, as both are in the corresponding clip reflecting overlapping audio sources or vague wording. This ambiguity results in low discrimination scores (e.g., VGGSound QA, Appendix Table~\ref{tab:dataset_issues_real}, row 1) and low feasibility.

False presuppositions, well-documented in text QA~\cite{DBLP:conf/acl/0001MZH23}, also appear in \aqa{} datasets such as FSD50K~\cite{DBLP:journals/taslp/FonsecaFPFS22}. These questions assume patterns not grounded in the audio, leading to inconsistent responses; correspondingly, IRT reveals low discrimination, indicating poor ability to differentiate annotator or model competence (Appendix Table~\ref{tab:dataset_issues_real_improved2}).

Weak grounding and shortcut learning are also prevalent, especially in caption-derived datasets like AudioCaps QA~\cite{DBLP:conf/naacl/KimKLK19} and AudioSet QA~\cite{DBLP:conf/icassp/GemmekeEFJLMPR17}. Questions such as \q{Is someone laughing?} exhibit low discrimination, as they can often be answered without audio, undermining evaluation validity (Appendix Table~\ref{tab:dataset_issues_real}).

Underspecified answer keys (e.g., \ans{dog}, \ans{puppy}, \ans{hound}) in datasets like MUSIC-21 QA~\cite{DBLP:journals/tismir/ChristodoulouGLJ25} and Clotho-\aqa{} introduce ambiguity, which IRT captures as reduced discrimination and noisy item estimates. Finally, overlapping audio sources (e.g., \mbox{VGGSound QA}, AudioCaps QA) increase ambiguity, producing items with high difficulty but low discrimination--- hard yet uninformative questions (Appendix Table~\ref{tab:merged_positioning_examples}).

%% file: 2025_arr_audio/sections/30-dataset.tex
\section{Audio Questions for Triva Experts}\label{sec:dataset}





We introduce a benchmark of human-authored audio questions designed by knowledgeable experts to test auditory and multimodal reasoning skills. This approach aligns with what \citet{DBLP:journals/csur/RogersGA23} call the ``probing'' paradigm and what \citet{DBLP:conf/emnlp/RodriguezB21} term the Manchester paradigm,
where questions are crafted explicitly to evaluate human-level understanding. While probing-style questions are common in text-only QA datasets such as~\citet{kabir-etal-2024-make} and TriviaQA~\cite{DBLP:conf/acl/JoshiCWZ17}, they remain underexplored in the audio domain.

We scrape questions from publicly available online audio tests, reflecting a broader practice across various communities-- such as educators, trivia enthusiasts, and researchers-- that release question sets to facilitate human study and benchmarking. This openness aligns with the Manchester paradigm’s emphasis on carefully crafted questions designed to probe deep understanding. However, unlike text-based questions, \aqa{} requires additional effort to segment continuous audio content into discrete, independently answerable questions that can be posed one at a time to humans or machines.
%
For instance, many examples have text instructions that apply to all of the following examples: e.g., \q{how many wheels do each of the following vehicles have}, followed by the sound that each of the vehicles makes.

In total, \name{} contains $9690$ audio-question pairs, providing a valuable resource for studying audio question answering.
Existing audio QA datasets differ substantially in scale, audio structure, and annotation philosophy, spanning large weakly-purposed corpora like VGGSound, human-curated benchmarks such as Clotho-AQA, and caption-derived QA sets like AudioCaps QA~(Table~\ref{tab:audita_scale_structure}). Despite their widespread use, these datasets are not systematically standardized: key properties such as question formulation, audio duration regimes, and annotation consistency vary significantly and are often incompletely reported. This lack of controlled design leads to confounded comparisons across methods and limits meaningful evaluation of auditory reasoning, motivating more rigorously constructed and diagnostically transparent benchmarks such as \name{}.

\input{2025_arr_audio/sections/tab_stat}

\begin{figure}[!tb]
    \centering
    \includegraphics[width=0.48\textwidth]{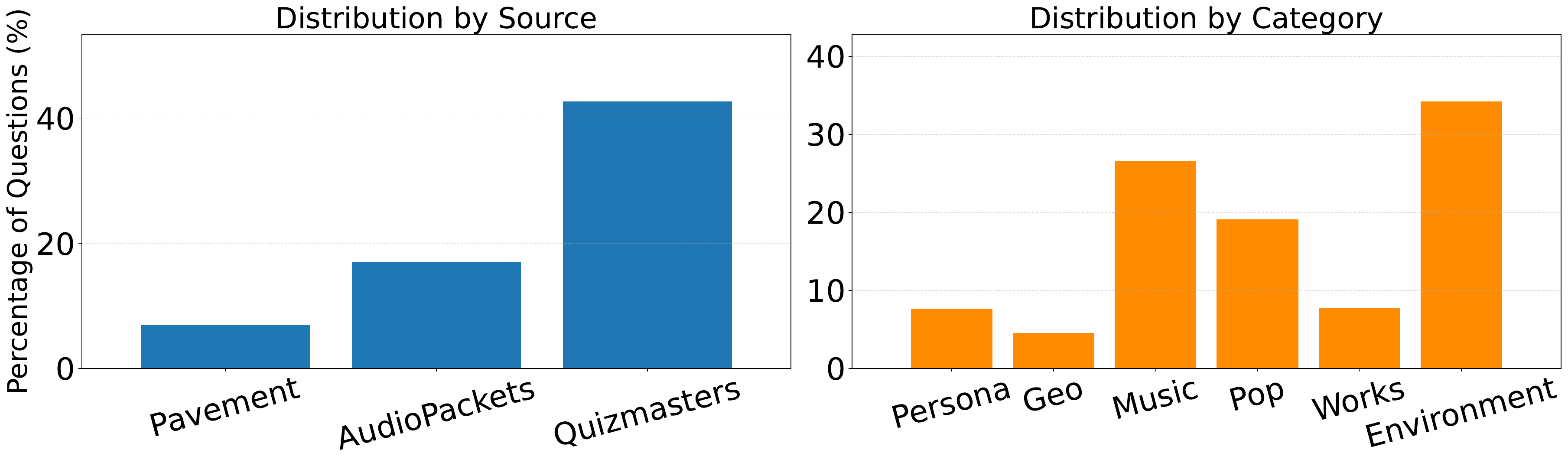}
    \caption{Distribution of questions across sources and categories in the dataset. The dataset exhibits diverse and rich coverage across both sources and categories, ensuring broad domain representation and supporting robust evaluation across varied audio question answering settings.}
    \label{fig:split}
\vspace{-0.4cm}
\end{figure}
\subsection{\name{} Data Sources}
We describe the sources used in \name{}, which span curated audio collections and structured trivia-style datasets.
\paragraph{Quizmasters Website}
The Quizmasters website~\cite{Quiizmasters} hosts curated collections of short audio clips organized by auditory skill category, originally created for pub quiz-style challenges, but without accompanying questions. These collections are designed to probe specific listening abilities, such as recognizing transformed audio (e.g., reversed or filtered signals) or identifying less common musical material. We convert these collections into an audio question answering format by attaching human-authored questions aligned with each category’s intended challenge. In addition to question–answer pairs, we retain clip-level metadata such as duration and sampling rate.

\paragraph{Audio Pyramidal Trivia and PAVEMENT}
Pyramidal question structure is established best practice for human question answering (HQA)~\cite{boyd2020question}, as it measures system knowledge by revealing clues incrementally from obscure to obvious. This principle extends naturally to audio: a pyramidal audio question might begin with an isolated instrument passage, progress through a fuller excerpt, and conclude with a recognizable theme-- rewarding systems that identify the answer from minimal acoustic context rather than only after highly identifying cues. Audio Pyramidal Trivia and PAVEMENT~\cite{DBLP:journals/corr/abs-1904-04792,pavement} both adopt this paradigm, structuring questions as sequences of progressively more informative audio clips. While earlier collections focus primarily on music (with some extensions to film and television), PAVEMENT~\cite{pavement} adds human-written competitive questions and richer annotations including acceptable answers and clarifications. Conventional QA benchmarks, by contrast, present questions as atomic, non-incremental units, leaving this evaluative dimension unaddressed. By retaining PAVEMENT's pyramidal structure and segmenting each question into individual audio question pairs, our evaluation captures a difficulty axis that flat CQA formats cannot.

To ensure correctness and consistency, we perform dataset cleaning and normalization, including verifying audio–question alignment and standardizing formatting. Figure~\ref{fig:split} and Appendix Table~\ref{tab:audita_sources} present the distribution of questions by source.

\subsection{Data Preparation}
\name{} is constructed through a three-stage pipeline:~alignment, normalization, and categorization to produce consistent and human-readable audio-question-answer triples for evaluation. The process ensures that questions are correctly paired with their corresponding audio clips, removes formatting artifacts, and standardizes answer representations to reduce variability across sources. Details of the scraping, cleaning, and normalization procedures are provided in Appendix~\ref{app:data_processing}.

\paragraph{Categorization and consolidation.}



Because our dataset and audio QA benchmarks more broadly span diverse auditory phenomena and reasoning skills, we organize questions into semantic categories. This structuring allows participants to answer queries relevant to their expertise, making the task more approachable and efficient for skilled annotators. It also enables fine-grained analysis of model behavior across domains (e.g., music recognition versus environmental sound reasoning).

We assign each question to a hierarchical taxonomy of six high-level categories and 26 subcategories using GPT-4o-mini. For usability in the human-centered QA task, we collapse this hierarchy into six interpretable categories exposed to participants: \textit{Cultural Geography in Sound}, \textit{Name The Music: Songs, Artists \& Composers}, \textit{Who's Who? Name That Persona}, \textit{Elements of Musical Works}, \textit{Pop Culture and Media}, and \textit{Environmental and Acoustic Sound Recognition}. This structure is applied uniformly across all data sources (Figure~\ref{fig:split} and Appendix Table~\ref{tab:audita_categories}).

\paragraph{MCQ generation}
We create MCQ variants with one correct answer and three AI-generated distractors validated by human annotators. The distractors are designed as plausible alternatives, sharing surface-level cues with the correct answer while differing in the key auditory evidence required for identification. MCQ variants help reveal both accuracy patterns and question quality, particularly in settings where models are not performing well and free-form responses may be influenced by language priors or generation artifacts. By requiring selection among closely related options, MCQs encourage discrimination rather than free-form generation or binary decisions, making them useful for diagnosing whether errors arise from model limitations or from the discriminative strength of the question.

For each question, we first determine the semantic type of the correct answer (e.g., musical artist, actor/actress, etc.), and then create three distractors that matched key attributes of the gold answer to ensure plausibility. For instance: If the correct answer was a \textbf{musical artist}, distractors match gender, genre, and era; be real existing artists; not be the same individual or an alias; and not violate entity type constraints (e.g., no band if the correct answer is a solo artist). If the correct answer was an \textbf{actor/actress}, distractors match gender, era, profession, and accent; be real individuals; and not be aliases of the correct answer.

Each row was generated independently to avoid cross-item leakage or systematic reuse patterns. \mbox{After} initial creation, a second author independently reviewed all distractors to verify that constraints were satisfied; all entities were valid and distinct; and no option was trivially eliminable.

\subsection{Dataset Composition}
\name{} is constructed from two sources: (i) original human-authored audio clips with associated questions, and (ii) external benchmark datasets. In total, it contains $9690$ questions, including $6460$ human-authored questions and $3230$ questions from external benchmarks. The human-authored portion is built on curated audio clips sourced from \textit{Pavements}, \textit{Audio-Packets}, and \textit{Quizmasters}. This includes $2322$ pyramidal-style questions ($673$ from \textit{Pavements} and $1649$ from \textit{Audio-Packets}) and $4138$ trivia-style questions from \textit{Quizmasters}. All human-authored questions are closed-ended with discrete, verifiable answers.

The external portion comprises $2907$ questions from Open\aqa{} (90\%) and $323$ questions from Clotho\aqa{} (10\%)~\cite{ltu,clothoaqa}. The external set contains $1205$ closed-ended questions ($882$ from Open\aqa{} and $323$ from ClothoAQA) and $2025$ open-ended questions (all from OpenAQA) that require semantic evaluation rather than exact string matching. For OpenAQA, we apply the original filtering procedure from~\citet{DBLP:conf/iclr/0001LLKG24}, removing $18.65\%$ of instances where the generation pipeline itself flags the question as unanswerable (e.g., responses containing “cannot be determined” or “unclear”), leaving only questions with committed answers~(details in Appendix Section~\ref{app:dataset-positioning}).

We include external datasets not as core evaluation targets but as \emph{reference points} to contextualize the behavior of our human-authored data. While OpenAQA and ClothoAQA are widely used audio QA benchmarks, they lack human validation, making it difficult to directly assess how close models are to human-level performance; including human responses enables this direct comparison. These datasets also provide an important contrast in difficulty and structure: external questions yield higher human accuracy and smaller human-model gaps, reflecting reliance on short, perceptual, or caption-derived cues. In contrast, our human-authored questions are explicitly designed to be more challenging and reasoning-intensive.

External datasets play a complementary role rather than serving as the primary benchmark. All analyses are disaggregated by source, and our main claims about human–model gaps focus on the human-authored subsets. ClothoAQA, despite structural limitations noted in Section~\ref{sec:motivation}, is included as an independently crowd-sourced dataset that provides a complementary human-annotated reference point beyond OpenAQA. Overall, the external portion serves to \emph{anchor} our evaluation-- enabling comparability with prior work while highlighting limitations of existing benchmarks relative to the reasoning demands of \name{}.

Across the dataset, in the human evaluation described in the next section, we collect $1517$ human guesses, individual answer attempts by participants, providing a reliable set of judgments to benchmark model performance. Our evaluation framework leverages these human guesses to jointly model question properties and participant abilities.

%% file: 2025_arr_audio/sections/tab_stat.tex
\begin{table}[t]
\centering
\tiny
\resizebox{\columnwidth}{!}{
\begin{tabular}{lrrrrr}
\toprule
Dataset & Total Qs & Unique Clips & Avg Q Len & Avg Dur (s) & Dur Range (s) \\
\midrule
VGGSound      & $\sim$200K & $\sim$200K & 12    & 10     & 10 \\
Clotho-AQA    & 11946     & 1991      & 14    & $\sim$22 & 15--30 \\
AudioCaps QA  & $\sim$50K  & $\sim$50K  & $\sim$10 & $\sim$5  & 0.5--10 \\
\name{}       & 9690      & 8713      & 12.47 & 36.98  & 0.42--478.33 \\
\bottomrule
\end{tabular}
}
\caption{Summary statistics describing the scale and structure of the \name{} benchmark compared to other \aqa{} datasets.}
\label{tab:audita_scale_structure}
\vspace{-0.3cm}
\end{table}

%% file: 2025_arr_audio/sections/40-experiments.tex
\section{How hard is \name{} for Humans and Computers}\label{sec:experiment}


We evaluate the proposed dataset using both state-of-the-art audio-language models and human participants to characterize its difficulty and suitability for auditory reasoning. We collect both free-form and multiple-choice (MCQ) responses under identical input conditions, enabling complementary assessment of generative and decision-based reasoning.


\subsection{Human Evaluation}
To ground performance in human capability and better understand the intrinsic difficulty of \name{}, we recruit participants from online quiz and trivia communities, who regularly engage with general-knowledge and audio-based trivia content. This provides a population with strong familiarity with trivia-style reasoning while remaining non-expert in the specific benchmark content.

We ask participants to evaluate the same audio clips and questions under identical conditions, without access to transcripts or external resources, ensuring that responses reflect purely auditory understanding. We elicit responses in both multiple-choice and free-response settings. Full instruction are provided in Appendix Section~\ref{human}.
\paragraph{Answer Evaluation}
Free-form answers are evaluated using the PEDANTS~\cite{DBLP:conf/emnlp/LiMNLB24} framework for semantic equivalence, allowing for variation in phrasing while preserving correctness. In contrast, MCQ responses are evaluated through selection among predefined answer options, providing a controlled setting for discriminative reasoning~(Appendix Section~\ref{answer_format}).
\subsection{Models}

We evaluate 18 models, including both open-source audio-multimodal systems and proprietary large models (GPT-4o and Gemini 2.5 Pro) to provide coverage across capability scales. The open-source models consist of 16 systems with mid-scale language backbones (approximately 4B–13B parameters), grouped into three categories. \textbf{Omnimodal models} (6) support unified understanding across text, audio, vision, and video with both text and speech generation. \textbf{Audio–language models} (4) focus on audio understanding with text-only outputs, while \textbf{speech-capable models} (6) emphasize speech recognition and generation, including both speech-first and modular architectures. GPT-4o and Gemini 2.5 Pro are included as representative state-of-the-art proprietary systems for comparison against recent large-scale multimodal models.

This taxonomy reflects the diversity of \name{} questions: some require speech or lyric understanding, others test purely acoustic reasoning over music or environmental sounds, and some benefit from broader multimodal context. Evaluating across capability groups enables analysis of whether failures are systematic or capability-specific, analogous to how humans identify songs via lyrics or melody. All models are evaluated using publicly released checkpoints (for open-source systems) and API-based inference (for proprietary models), with recommended settings and no task-specific fine-tuning. Full model details are provided in Appendix~\ref{app:model-details}.

\subsection{Input Representation}

To fairly evaluate both humans and computational models on our audio question answering task, we ensure consistent presentation of audio and questions. Each example consists of an audio clip paired with a natural-language question, presented verbatim without templating or normalization. For human participants, audio clips are provided in standard formats (.mp3, .wav) to ensure consistent playback across devices.

For computational models, audio inputs are uniformly preprocessed: audio is converted to mono and resampled to the sample rate expected by each model. Depending on the model’s input requirements, we supply either raw waveform audio or alternative audio representations such as log-mel spectrograms or codec tokens, following each model’s prescribed preprocessing pipeline.


%% file: 2025_arr_audio/sections/50-analysis.tex
\section{Human--Model Performance Gaps with Accuracy and IRT}\label{sec:analysis}

Humans have a substantial and consistent lead over state-of-the-art audio-language models on \name{} across both free-form and multiple-choice settings (human: \HUMANACCURACY\% vs.~models: \MODELACCURACY\% in free-form answers; human: \HUMANMCQACCURACY\% vs.~models:~15.65\% in MCQ). However, accuracy alone obscures important structure in QA benchmarks.
\input{2025_arr_audio/sections/model_table}
\input{2025_arr_audio/sections/model_detail}

\begin{figure}
    \centering
    \includegraphics[width=0.5\textwidth]{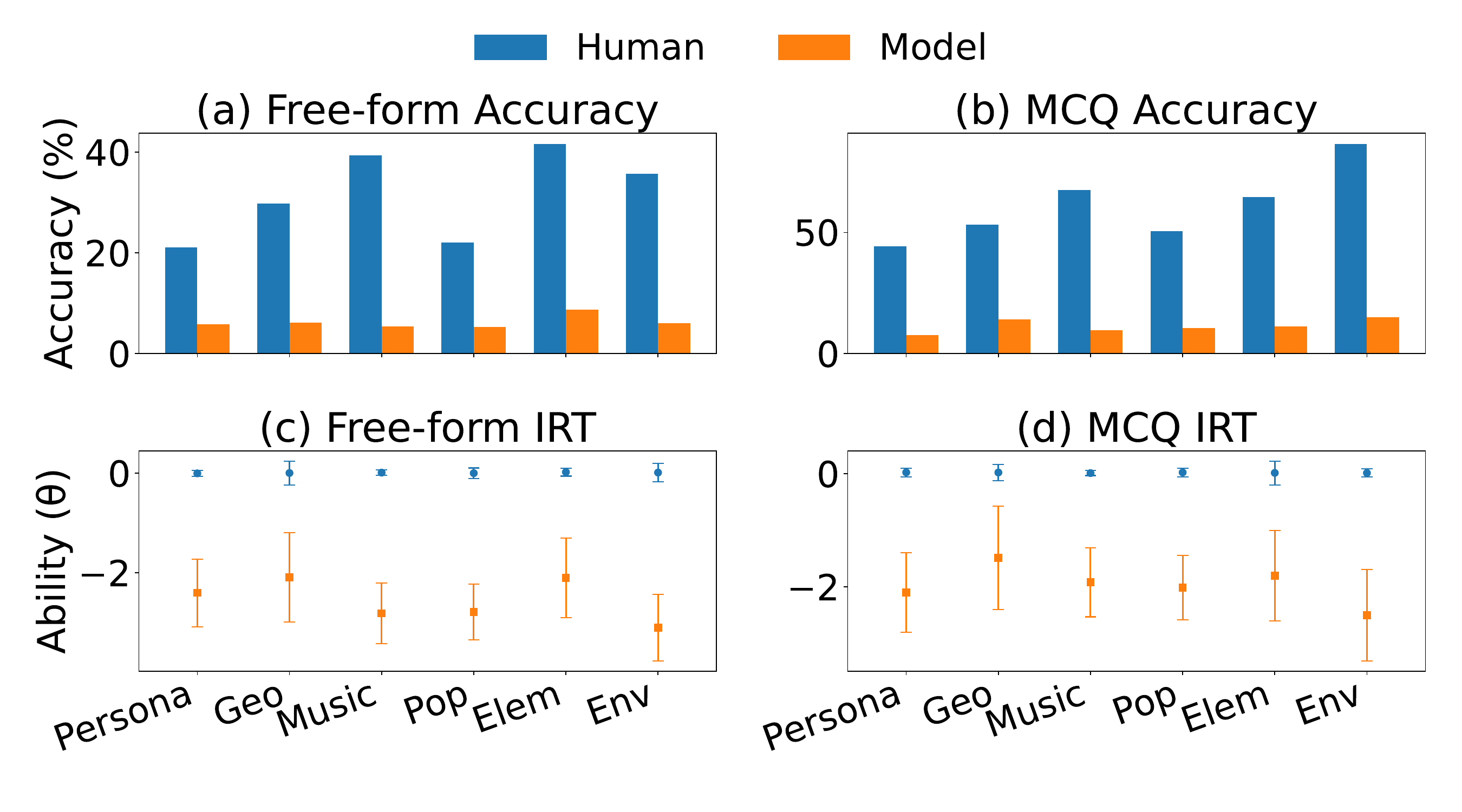}
    \caption{Category-level accuracy and average item difficulty for humans and models on free-form and multiple-choice questions. Bars show accuracy; points show mean IRT difficulty $\pm$ one standard deviation. Higher difficulty correlates with lower accuracy for both, but models’ performance declines more sharply, revealing systematic category-specific gaps.}
    \label{fig:category}
\vspace{-0.3cm}
\end{figure}

While these accuracy gaps clearly indicate strong human advantages, they do not capture the quality or diagnostic value of individual questions in \name{}. In particular, questions vary substantially in how effectively they distinguish between high- and low-ability respondents.

We use \textbf{Item Response Theory (IRT)} as a principled framework for analyzing human and model performance and diagnosing evaluation quality~\cite{baker2004item,embretson2013item}. Widely used in standardized testing and leaderboard construction, IRT enables comparison across heterogeneous populations by normalizing for question difficulty~\cite{rodriguez-boyd-graber-2021-evaluation}. This is particularly important in \name{}, where items vary substantially in acoustic complexity and reasoning demands. By jointly modeling difficulty and discrimination, IRT supports analysis beyond aggregate accuracy and identifies which questions are most informative. For example, in \name{} high-discrimination items include entity identification from audio clips, while low-discrimination items often involve underspecified prompts such as language identification or continuation-based queries (Appendix Table~\ref{tab:irt_discrimination_examples}).

Table~\ref{tab:summary_stats} and Table~\ref{tab:model_leaderboard} report accuracy alongside IRT ability estimates for humans and models. In addition to measuring performance, IRT highlights poorly diagnostic items: low-discrimination questions fail to separate high- and low-ability respondents, while extremely easy or difficult items provide limited signal for comparison. Humans substantially outperform all models under both metrics, and system rankings remain broadly consistent across accuracy and ability.

Both humans and models perform well on easier questions, while the performance gap widens with increasing difficulty, with humans remaining more robust on harder items. Easier questions typically contain clear audio cues, familiar sounds, or require limited temporal integration, corresponding to lower IRT difficulty ($b$) values. Earlier benchmarks tend to exhibit a narrower difficulty range, where models achieve higher accuracy and show less variation in estimated ability ($\theta$). In contrast, \name{} spans a broader difficulty spectrum, producing larger and more consistent human–model gaps and better exposing model limitations. Additional item- and dataset-level breakdowns are provided in Appendix Figure~\ref{fig:item} and Table~\ref{tab:text_mcq_split}.
\begin{figure}[!tb]
    \centering
    \includegraphics[height=0.15\textwidth,width=0.3\textwidth]{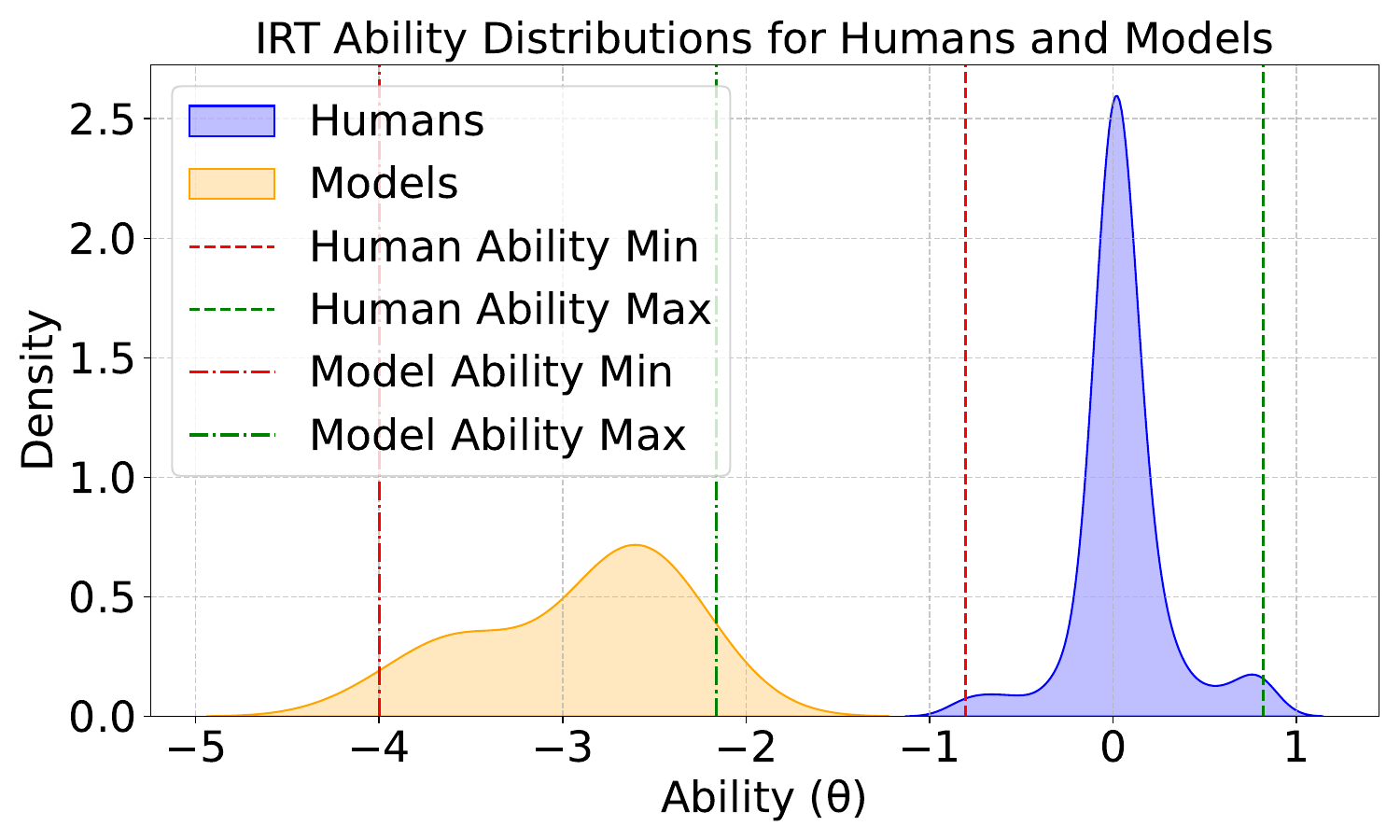}

    \caption{Distributions of IRT ability ($\theta$) for humans (blue) and models (orange) are shown on a shared scale. Kernel density estimates highlight that humans cluster at higher $\theta$, with dashed lines indicating each group’s range, revealing a clear latent ability gap despite model variability.}
    \label{fig:ability}
\vspace{-0.4cm}
\end{figure}

To localize this gap, Figure~\ref{fig:category} presents category-level accuracy alongside mean item difficulty. Categories involving music and environmental sound consistently show strong human performance but weak model performance. For example, in Elements of Musical Works, humans reach 41.6\% (free-form) and 64.8\% (MCQ), while models remain low at 8.7\% and 11.3\% (IRT ability estimate $\theta$:0.03 vs.\ -2.1). Similar gaps appear in Name The Music (39.4\% / 67.6\% vs. 5.4\% / 9.7\%) and Environmental Sound Recognition (86.7\% vs. 14.9\% MCQ). Overall, consistent gaps across categories indicate that failures stem from intrinsic auditory reasoning difficulty rather than annotation or evaluation noise. Figure~\ref{fig:ability} shows the IRT ability distributions for humans and models on a shared scale. Models concentrate in a narrower, substantially lower ability range than humans, indicating reduced robustness as item difficulty increases.




Category-level analysis shows that environmental sounds and complex music remain particularly challenging for models, while humans remain strong across these domains. Models often confuse acoustically similar clips or fail to capture long-range temporal cues, performing relatively better in MCQ settings but still showing substantial gaps. Overall, IRT-based analysis confirms that models struggle most on high-difficulty items, highlighting limitations in robust auditory reasoning.

\begin{table}[t]
\centering
\tiny
\resizebox{0.8\columnwidth}{!}{
\begin{tabular}{lcr}
\toprule
\textbf{Setting} & \textbf{Input Modalities} & \textbf{Accuracy} \\
\midrule
Text-only & Question only (no audio) & 0.0001\% \\
Transcript-only & Transcript (no audio) & 4.26\% \\
Multimodal & Audio + text & 8.86\% \\
Text-only MCQ & Question + options (no audio) & 1.29\% \\
Transcript-only MCQ & Transcript + options (no audio) & 7.05\% \\
Multimodal MCQ & Audio + text + options& 15.65\% \\
\bottomrule
\end{tabular}}
\caption{Near-zero and lower performance in text-only and transcript-only settings and gains from multimodal input confirm that \name{} cannot be solved using textual priors alone and requires audio understanding.}
\label{tab:baseline}
\vspace{-0.45cm}
\end{table}
\paragraph{Baseline Analysis and MCQ Performance}
To isolate the contribution of different modalities in \textsc{AUDITA}, we evaluate text-only, transcript-only, and multimodal settings under both free-form and MCQ formats (Table~\ref{tab:baseline}).
Text-only settings achieve near-zero accuracy, confirming that the benchmark cannot be solved using linguistic priors alone. Even with transcripts, performance remains low (4.26\%), as many examples contain minimal or no spoken content (e.g., music, environmental sounds, speaker identity). In contrast, multimodal inputs improve performance relative to text-only variants, demonstrating that non-zero accuracy depends on access to acoustic information.

In the MCQ setting, accuracy (15.65\%) falls below the 25\% chance level. This does not indicate random guessing; instead, it reflects systematic miscalibration, where models consistently prefer plausible but incorrect options due to fine-grained acoustic distinctions between distractors~\cite{zhao2021calibrate,wang2025llms}. Similar behavior has been observed in prior multimodal evaluation settings, where models rely on spurious correlations or superficial cues that misalign with the underlying evidence~\cite{Kavumba2022prompt,wumagic}. These results show that \name{} evaluates genuinely audio-grounded understanding rather than text-based reasoning.
\paragraph{Interpreting human accuracy}
\name{} consists of open-ended audio trivia with a large effective answer space, making chance performance in free response effectively zero. In multiple-choice, chance accuracy is $1/K$ (e.g., 25\% for $K$=4), yet humans substantially exceed this baseline under identical conditions. Free-form-answer scoring is intentionally strict; many items admit plausible near-misses (e.g., confusing franchise installments or covers vs.\ originals)--so even moderate free-form response accuracy reflects meaningful auditory reasoning rather than guessing, consistent with human baselines reported by MMAU~\cite{sakshi2024mmaumassivemultitaskaudio}.

%% file: 2025_arr_audio/sections/model_table.tex
\begin{table}[!t]
\centering
\tiny
\setlength{\tabcolsep}{3pt}
\renewcommand{\arraystretch}{0.9}
\resizebox{\columnwidth}{!}{
\begin{tabular}{lrrrrrrrrr}
\toprule
&\multicolumn{4}{c}{Free-from}&\multicolumn{4}{c}{MCQ}\\
\multicolumn{1}{c}{System}
& \multicolumn{1}{c}{Accuracy}
& \multicolumn{1}{c}{$\theta$}
& \multicolumn{1}{c}{SD}
& \multicolumn{1}{c}{95\% CI}
& \multicolumn{1}{c}{Accuracy}
& \multicolumn{1}{c}{$\theta$}
& \multicolumn{1}{c}{SD}
& \multicolumn{1}{c}{95\% CI} \\
\midrule

Humans
& {\HUMANACCURACY}
& {0.05}
& 0.26
& {[-0.001, 0.101]}
& {\HUMANMCQACCURACY}
& {0.08} & 0.25
& {[0.055, 0.105]} \\

Models (avg)
& {\MODELACCURACY}
& {-2.91}
& 0.55
& {[-3.097, -2.723]}
& 15.65
& {-2.45} & 0.54
& {[-2.636, -2.264]} \\

\bottomrule
\end{tabular}}
\caption{Human vs model aggregate performance with IRT-based ability estimates and uncertainty intervals. Humans substantially outperform models across both text and MCQ settings, with a large and consistent gap in estimated ability ($\theta$), indicating a strong human–model performance disparity beyond raw accuracy.}
\label{tab:summary_stats}
\end{table}

%% file: 2025_arr_audio/sections/model_detail.tex
\begin{table}[!t]
\centering
\tiny
\setlength{\tabcolsep}{3pt}
\renewcommand{\arraystretch}{0.9}
\resizebox{\columnwidth}{!}{
\begin{tabular}{lrrrrrr}
\toprule
&\multicolumn{3}{c}{Free-from}&\multicolumn{3}{c}{MCQ}\\
{Model}
& {Acc.}
& {$\theta$}
& {Rank}
& {Acc.}
& {$\theta$}
& {Rank} \\
\midrule

Gemini 2.5 Pro~\cite{Comanici2025Gemini2P}
& 17.02 & -1.66 & 1 & 39.18 & -1.40 & 1 \\

GPT-4o~\cite{Hurst2024GPT4oSC}
& 14.87 & -1.97 & 2 & 34.89 & -1.62 & 2 \\

Qwen2.5-Omni~\cite{DBLP:journals/corr/abs-2503-20215}
& 10.01 & -2.16 & 3 & 21.02 & -1.76 & 4 \\

AudioGPT~\cite{DBLP:conf/aaai/HuangLYSCYWHHLR24}
& 8.98 & -2.31 & 4 & 23.49 & -1.61 & 3 \\

OpenOmni~\cite{DBLP:journals/corr/abs-2501-04561}
& 7.99 & -2.45 & 5 & 16.01 & -2.06 & 6 \\

Audio-Flamingo~\cite{DBLP:conf/icml/KongGBPVC24}
& 7.77 & -2.49 & 6 & 13.99 & -2.11 & 8 \\

Phi-4-Multimodal~\cite{DBLP:journals/corr/abs-2503-01743}
& 7.49 & -2.54 & 7 & 15.73 & -2.09 & 7 \\

Qwen3-Omni~\cite{DBLP:journals/corr/abs-2509-17765}
& 6.87 & -2.62 & 8 & 18.89 & -2.04 & 5 \\

LTU-AS~\cite{DBLP:conf/asru/GongLLKG23}
& 6.69 & -2.62 & 9 & 13.81 & -2.11 & 9 \\

Qwen-2 Audio~\cite{DBLP:journals/corr/abs-2407-10759}
& 6.53 & -2.70 & 10 & 13.72 & -2.31 & 10 \\

Baichuan-Omni-1.5~\cite{DBLP:journals/corr/abs-2501-15368}
& 6.49 & -2.71 & 11 & 13.63 & -2.43 & 11 \\

VITA-1.5~\cite{DBLP:journals/corr/abs-2501-01957}
& 5.76 & -2.81 & 12 & 12.59 & -2.71 & 12 \\

Mini-Omni2~\cite{DBLP:journals/corr/abs-2410-11190}
& 4.85 & -3.03 & 13 & 8.74 & -2.78 & 14 \\

SpeechGPT~\cite{DBLP:conf/emnlp/ZhangLZZWZQ23}
& 3.99 & -3.28 & 14 & 12.38 & -2.71 & 13 \\

SALMONN-2~\cite{DBLP:journals/corr/abs-2506-15220}
& 2.77 & -3.60 & 15 & 5.68 & -3.31 & 17 \\

SALMONN-2+~\cite{DBLP:journals/corr/abs-2506-15220}
& 2.62 & -3.61 & 16 & 6.28 & -2.89 & 16 \\

MU-LLaMA~\cite{DBLP:conf/icassp/LiuHSS24}
& 2.19 & -3.61 & 17 & 7.48 & -2.81 & 15 \\

SALMONN~\cite{DBLP:conf/iclr/TangYSC000M024}
& 1.99 & -4.00 & 18 & 4.18 & -3.40 & 18 \\

\bottomrule
\end{tabular}}

\caption{Full model comparison on text and MCQ settings using accuracy and IRT-based ability estimates. State-of-the-art proprietary models (e.g., Gemini 2.5 Pro and GPT-4o) lead across both settings, while open-source models exhibit a large performance spread, with consistent rankings across accuracy and IRT-based ability estimates.}
\label{tab:model_leaderboard}
\end{table}

%% file: 2025_arr_audio/sections/51-error.tex
\subsection{Factual Knowledge vs.\ Audio Understanding}

Some questions in \name{}-for example, \q{Name the artist who recorded this song} or \q{What film is this theme from?}- require both perceptual recognition and world knowledge. To disentangle these factors, we compare two model settings already available in our experiments: \textit{Question + Transcript} (Whisper output) and \textit{Question + Raw Audio} (see Appendix~\ref{app:factual-audio} for full details and examples).

This comparison allows us to interpret performance differences diagnostically. \mbox{Consider} \q{Name the songwriter} with the answer \ans{Joni Mitchell(River)}. Here, the transcript contains lines from the song, enabling correct identification, whereas the raw audio alone does not provide enough cues, representing a perceptual limitation. In contrast, \q{You are listening to the voice of a fictional character; what is this character's name?} with the answer \ans{Frank Spencer}, both transcript and audio fail, reflecting a knowledge limitation. Finally, in cases such as \q{Name the composer} with the answer \ans{Carl Orff}, the distinctive choruses and instrumental timbres in the audio allow the model to succeed, while the transcript contains no usable information; this highlights reliance on non-verbal acoustic cues.

Across all items, accuracy is substantially higher in the raw-audio condition (8.86\%) than in the transcript condition (4.26\%). If performance were primarily driven by textual or factual recall, transcript-based accuracy would match or exceed audio-based results. This gap indicates that transcripts often omit or distort critical acoustic information, making non-verbal audio cues essential for identification. This difference shows that failures cannot be explained solely by missing world knowledge, but reflect genuine limitations in audio understanding.

\subsection{Failure Case Analysis}
We analyze model failures in \name{} through three error modes: knowledge-based, perceptual, and audio-cue reasoning failures. Knowledge-based errors dominate at 78.23\%, while the remaining 21.77\% involve failures where auditory perception or acoustic cues play a decisive role. A key finding is that even when models have access to audio signals, they frequently fail on tasks requiring fine-grained auditory reasoning combined with cultural or entity-level knowledge. For example, music recognition and cultural media questions often require identifying composers, artists, or works from subtle acoustic signatures such as melody structure, timbre, orchestration, or vocal style. In \textit{Name The Music}, humans achieve 39.4\% (free-form) and 67.6\% (MCQ) accuracy, while models remain at 5.4\% and 9.7\%, highlighting a large gap in auditory grounding and entity recognition. Similarly, in \textit{Environmental and Acoustic Sound Recognition}, humans reach 35.7\% MCQ accuracy compared to 14.9\% for models, indicating that humans can reliably infer sound sources from acoustic structure, whereas models struggle to form stable perceptual representations of environmental sounds.

Perceptual and audio-cue errors also highlight limitations in how models exploit acoustic information. Some failures stem from weak use of discriminative audio structure, while others reflect difficulty capturing temporal patterns, texture, and non-speech cues absent from text. These cases show that success requires not only linguistic reasoning but also direct acoustic grounding and temporal pattern recognition in time-varying signals.

Knowledge-based failures remain the largest category, but they often co-occur with weak or underspecified acoustic evidence. Many questions require identifying culturally grounded entities such as operas, songs, or fictional characters, where success depends on linking auditory cues with prior world knowledge. When either component is weak, performance degrades sharply. These results suggest that errors in \name{} are not solely due to missing factual knowledge. Instead, they stem from failures to jointly integrate auditory perception (e.g., timbre, rhythm, and sound structure) with entity-level reasoning over cultural and factual knowledge. This interaction explains why models fail even with access to audio, highlighting a fundamental gap in auditory grounding compared to human listeners (details in Appendix Section~\ref{app:failure-analysis}).

%% file: 2025_arr_audio/sections/55-related_work.tex
\section{Related Work}\label{sec:related_work}
We review prior work in audio QA, multimodal and adversarial evaluation, and identify key limitations motivating \name{}.

\paragraph{\aqa{} datasets}
Audio Question Answering has evolved from synthetic benchmarks to more natural but still limited human-sourced datasets. Early works such as CLEAR~\cite{DBLP:conf/nips/LinSPR21} generate synthetic acoustic scenes with compositional questions, enabling controlled evaluation but limiting realism. DAQA~\cite{DBLP:journals/taslp/FayekJ20} similarly uses constrained sound compositions with yes/no or count-based questions, reducing reasoning to structured classification. More natural datasets such as ClothoAQA~\cite{DBLP:conf/icassp/DrossosLV20} and Music-AVQA~\cite{DBLP:conf/cvpr/LiWTXW022} introduce crowd-sourced questions over real audio or video. However, they still exhibit strong linguistic priors and template bias, allowing models to succeed with limited audio grounding and often overestimating performance (Appendix Table~\ref{tab:superhuman-model-performance}). The DCASE 2025 Audio QA challenge~\cite{DBLP:journals/corr/abs-2505-07365} expands domain coverage but remains largely multiple-choice and non-adversarial. Prior work further shows sensitivity to textual artifacts and shortcut cues in audio-language models~\cite{DBLP:journals/corr/abs-2508-15407}, suggesting current benchmarks do not reliably measure robust auditory reasoning.

\paragraph{Adversarial and multimodal QA}
Outside audio, adversarial QA exposes model weaknesses via human-in-the-loop construction of challenging examples. In vision-language QA, methods such as Adversarial VQA and Dynabench~\cite{DBLP:conf/nips/ShengSGMTGPK21,kiela-etal-2021-dynabench} iteratively generate harder questions via model feedback. Recent analyses show that many benchmarks fail to induce true adversarial behavior despite appearing challenging~\cite{sung-etal-2025-benchmark}. In text QA, adversarial datasets such as~\cite{wallace-etal-2019-universal} reduce shortcut reliance and improve robustness evaluation. Multimodal QA benchmarks including VQA~\cite{DBLP:conf/iccv/AntolALMBZP15}, CLEVR~\cite{DBLP:conf/cvpr/JohnsonHMFZG17}, GQA~\cite{DBLP:conf/cvpr/HudsonM19}, and MultimodalQA~\cite{DBLP:conf/iclr/TalmorYCLWAIHB21} address bias and compositionality via balancing and structured grounding. Video and audio-visual QA datasets like AVQA~\cite{DBLP:conf/mm/Yang0DCHJ022} and Music-AVQA~\cite{DBLP:conf/cvpr/LiWTXW022} extend these ideas temporally but still rely largely on templated or crowd-sourced questions (Appendix Section~\ref{app:related_work}).

\paragraph{Positioning relative to prior \aqa{} benchmarks}
Prior \aqa{} benchmarks focus on recognition-style tasks: closed-set event labeling (e.g., \ans{dog barking}, \ans{siren}), caption- or metadata-derived QA (e.g., \q{What instrument is playing?}), synthetic or templated mixtures, and speech-centric pipelines reducing audio QA to ASR followed by text QA. As summarized by MMAU~\cite{sakshi2024mmaumassivemultitaskaudio}, these settings test information extraction with limited reasoning. In contrast, \name{} targets a regime requiring of grounding audio to real-world entities and integrating long-range cues. Instead of identifying sounds, it asks questions such as \q{Which song is this clip from?}, \q{Which film does this score belong to?}, or \q{Which artist is associated with this segment?}. Answering these requires audio-to-referent grounding and reasoning over distributed acoustic cues (e.g., melody, timbre, context), shifting the task from surface recognition to entity-level inference.
\paragraph{Item Response Theory in NLP evaluation}
Item Response Theory (IRT), originally developed in psychometrics, has been increasingly adopted in NLP for evaluating datasets and models. \citet{lalor2016building,rodriguez-boyd-graber-2021-evaluation} show that modeling item difficulty and discrimination provides a more informative alternative to aggregate accuracy and enables comparison across heterogeneous examples and systems. IRT has also been used to scale evaluation using model-generated responses and to diagnose dataset quality by identifying uninformative or low-discrimination items \cite{lalor2020learning}.

%% file: 2025_arr_audio/sections/60-conclusion.tex
\section{Conclusion}\label{sec:conclusion}


We present \name{}, a benchmark of human-authored audio questions grounded in real-world settings, designed to evaluate genuine auditory reasoning. Our results show a clear and persistent gap between human and model performance, with IRT analysis revealing limitations in handling acoustic cues, temporal structure, and context. These findings suggest that scaling alone is insufficient. 

Future work should focus on improving audio-language alignment, strengthening entity-level reasoning, and enabling multi-step inference over complex audio inputs, potentially through retrieval-augmented approaches that incorporate external knowledge sources. More fine-grained diagnostic frameworks can further help isolate and address these failures.

%% file: 2025_arr_audio/sections/70-acknowledgements.tex
\section{Limitations}

Our evaluation targets out-of-the-box performance of locally runnable, open-checkpoint models in a mid-scale regime, and we do not claim to cover the full space of proprietary or cloud-only systems. Larger closed models may improve absolute accuracy, but the scale question is discussed directly in Appendix~\ref{app:scale_validity}, including why scale alone is unlikely to close the gap we observe on a difficult, human-authored benchmark. We also do not run full scaling sweeps across cloud-scale systems because evaluating 9{,}690 items with audio inputs would impose substantial cost.

We intentionally evaluate end-to-end model behavior without external tool augmentation. In particular, we do not benchmark pipelines that add ASR plus retrieval, music fingerprinting, database lookups, or web search. These systems are relevant in practice and may reduce errors on referent-linking items, but they measure a different capability than the audio-grounded reasoning we aim to isolate. Relatedly, we evaluate all systems in an audio+question to text-answer setting for consistency, even when a model can generate speech, which may understate the value of speech-first interaction designs in real deployments.

Our psychometric analysis depends on the breadth and quality of human responses. While IRT helps separate item difficulty from responder ability, it can still be affected by annotator variability and by items where the audio is noisy, overlapping, or underspecified. Human participants are also not uniformly “trivia experts,” so aggregate human accuracy should be interpreted as a baseline rather than a ceiling. Finally, although \name{} is deliberately sourced from real-world domains, it is still shaped by the distribution of publicly available audio trivia and by English-centric question writing, which may limit generalization to other languages, accents, and niche audio domains.

\name{} contains short audio clips (averaging 37 seconds) from publicly available trivia sources, following practices established by multimedia QA benchmarks such as TVQA \citep{lei2018tvqa}. We will release stable metadata and acquisition scripts to support reproducibility.

\section{Ethical Considerations}
Human evaluation was conducted under Institutional Review Board (IRB) approval and informed consent. We collected participant email addresses solely to deliver remuneration. Emails are stored securely, are not used for analysis, and are not linked to response data in the released benchmark or reported results. Aside from compensation logistics, we do not collect additional personally identifying information, and we analyze results in anonymized form. The audio content used in \name{} is sourced from publicly available datasets or permissively licensed materials, and is used for research purposes. We do not redistribute restricted content beyond derived annotations required for benchmarking.

\section*{Acknowledgement}

This work is supported by TRAILS (Institute for Trustworthy AI in Law \& Society) (CNS-2150382). We thank the University of Maryland Institute for Advanced Computer Studies (UMIACS) for continuous support and computational resources. We also thank the reviewers for their valuable comments and suggestions, which helped improve the clarity and quality of this work. We are grateful to all participants who contributed to the human evaluation. The following participants explicitly consented to being acknowledged by name: Daniel Kim, Drew Scheeler, Eve Nuria Fleisig, Forrest Weintraub, Hemanth Nandakumar, Jason Christopher, Mohammed Afaan Mohammed Arif Ansari, Nathan Zhao, Nishant Balepur, Raymond Kimball, Sara DelVillano, and Stefany Meyer. Their annotations and responses were essential for establishing reliable human performance benchmarks and validating the quality and difficulty of the dataset. We especially appreciate the time and effort they invested in engaging with challenging audio-based questions. We also acknowledge the authors of prior datasets and resources used in this work for making their data and tools publicly available.

%% file: 2025_arr_audio/sections/appendix-related_work.tex
\appendix
This appendix provides detailed documentation of the data sources, processing pipeline, evaluation settings, and extended analyses used to construct and study \name{}. Section~\ref{app:data_processing} describes data collection and processing, including source-specific details for Quizmasters and PAVEMENT, as well as extraction, alignment, and normalization procedures. It also includes dataset statistics (Tables~\ref{tab:audita_sources} and~\ref{tab:audita_categories}) and qualitative analyses of existing audio QA datasets (Tables~\ref{tab:merged_positioning_examples}, \ref{tab:dataset_issues_real}, and \ref{tab:dataset_issues_real_improved2}).

Section~\ref{answer_format} outlines the evaluation settings, including free-response scoring using PEDANTS and multiple-choice evaluation. Section~\ref{app:model-details} details the evaluated models, their architectures, training paradigms, and capability groupings. Section~\ref{app:related_work} situates \name{} within prior work on audio QA, multimodal QA, and adversarial evaluation.

Section~\ref{app:scale_validity} examines the role of model scale and validates that our conclusions are not artifacts of model size alone. The main results and diagnostic analyses are presented in Section~\ref{app:mcq-analysis}, including MCQ behavior, human agreement, task difficulty, and IRT-based evaluation (Section~\ref{IRT}). Section~\ref{app:factual-audio} and Section~\ref{app:failure-analysis} further analyze model failures across modalities, distinguishing perceptual, knowledge-based, and audio-driven errors.

Finally, Section~\ref{human} documents the human evaluation protocol and participant setup, with additional details in Sections~\ref{ins} and~\ref{participant_list}. Together, these sections provide a comprehensive view of dataset construction, evaluation methodology, and the underlying factors driving model performance on \name{}.
\section{IRT Model Specification}\label{IRT}

\textbf{Item Response Theory (IRT)} is an increasingly common method for discovering gaps between human and machine ability and for identifying problematic examples~\cite{baker2004item,embretson2013item}.

We adopt the standard \textbf{two-parameter logistic (2PL) IRT model}. In this framework, each model (or human group) is treated as a ``respondent'' with ability parameter $\theta$, and each question is characterized by two parameters:

\begin{enumerate}
    \item \textbf{Difficulty} $b$
    \item \textbf{Discrimination} $a$
\end{enumerate}

The probability that a respondent with ability $\theta$ correctly answers item $i$ is given by

\begin{equation}
P(\text{correct} \mid \theta) = \sigma(a_i(\theta - b_i))
\end{equation}

where $\sigma(\cdot)$ denotes the logistic function.

Intuitively:

\begin{itemize}
    \item $\theta$ (\textbf{ability}) represents the overall skill level on the benchmark. Higher $\theta$ indicates a higher probability of answering difficult items correctly.
    \item $b$ (\textbf{difficulty}) represents how challenging a question is. Items with larger $b$ require higher ability to achieve 50\% correctness~\cite{embretson2013item}.
    \item $a$ (\textbf{discrimination}) measures how sharply an item distinguishes between high- and low-ability respondents. Higher $a$ means the item better separates strong from weak systems.
\end{itemize}

The raw parameters are not interpretable in isolation, but they enable comparison \textbf{between agents}. For example, a value such as $\theta = -2.91$ does not correspond to ``$-291\%$ accuracy'' or any direct percentage. Rather, it indicates that the model's ability is substantially below the dataset's average difficulty level (which is centered near $0$). Concretely, when $\theta \ll b$, the logistic function yields a low probability of correctness across many items, especially those with moderate or high discrimination. Thus, strongly negative $\theta$ reflects consistent failure even on moderately difficult questions, not just lower raw accuracy.

The numbers reported in \textbf{Table~4} are obtained by fitting the 2PL model to the binary correctness matrix (respondent $\times$ item) via maximum likelihood estimation. Item parameters $(a_i, b_i)$ and respondent abilities $(\theta_j)$ are estimated jointly under standard identifiability constraints (e.g., fixing the mean ability to $0$). Once fitted, each model's $\theta$ is directly obtained from the estimated parameters.

\section{Data Collection and Processing Details}\label{app:data_processing}

\paragraph{Quizmasters Website} The Quizmasters website~\footnote{\url{https://www.thequizmasters.biz/}} 
publishes standalone audio clips grouped under categorical umbrellas, but without associated questions. These clips range from 3 to 40 seconds in length and are organized into categories that either apply audio transformations (e.g., reversal) or present challenging identification tasks involving less popular material. Since clips within a category share common properties, we assign handwritten questions appropriate to the category format. For example, a clip from the National Anthems category may be paired with the question \q{What country is this national anthem from?} Some collections also include a corresponding reveal clip containing the original, untransformed audio; because these clips do not exhibit the intended transformation, we instead assign identification-style 
questions such as ``Name the artist who recorded this song.'' In addition to question–answer pairs, we store clip-level metadata including sampling rate, duration, and other audio attributes.

\paragraph{PAVEMENT Processing}
Each dataset entry includes a \textit{notes} field that preserves additional information from the original QuizBowl questions, such as alternative acceptable answers or clarifying remarks. These annotations are retained as part of the source material. In the PAVEMENT subset, QuizBowl questions are split into individual clues for model evaluation.

Some questions refer to shared attributes across clues, such as \q{What is the common number in the titles of these songs?} or \q{What is the common profession mentioned in the titles of these songs?} While some tournaments, such as SoundTrack, provide clip descriptions in their GitHub repositories, this practice is inconsistent and absent in PAVEMENT.

Our processing of PAVEMENT focused on data integrity. Earlier scraped versions contained mismatches between audio clips and questions, ordering errors, and formatting issues. We therefore re-scraped the dataset to ensure correct question–answer alignment and applied normalization steps to make outputs human- and model-ready, including resolving Unicode issues, inconsistent notation, and formatting irregularities. 
\subsection*{Data Preparation}
We prepare AUDITA in three stages: (i) extraction and alignment, (ii) normalization for evaluation, and (iii) categorization for analysis. The goal is not to change the underlying questions, but to make the resulting audio--question--answer triples consistent, human-readable, and robust to evaluation artifacts.

\paragraph{Extraction and alignment.}\label{extraction and alignemnt}

For GitHub-hosted QuizBowl-style sources, we extract question prompts and answerlines from the provided materials and align them with audio files using source-specific directory conventions and indexing. For \textit{Pavements}, we correct earlier indexing mismatches by enforcing consistent clip identifiers and alignment rules between prompts, answerlines, and audio files. Table~\ref{tab:audita_categories} reports the resulting category distribution.

\paragraph{External benchmark preparation.}
To provide a concrete comparison point to prior audio question answering resources, we include questions from OpenAQA and ClothoAQA. OpenAQA is largely generated from captions and metadata as part of LTU's OpenAQA-5M pipeline, while ClothoAQA is crowdsourced, which offers a small human-written counterpoint to caption-derived questions~\cite{ltu,clothoaqa}. We do not rewrite the benchmark questions or answers to make them more human-friendly. Instead, we apply only minimal preprocessing needed for evaluation consistency.

For OpenAQA, we run the filtering utilities released with LTU to remove unanswerable or hallucinated question--answer pairs~\cite{ltu}. In our snapshot, this removes 18.65\% of candidate items. We then sample proportionally across OpenAQA's constituent source datasets to preserve its original mixture.


\paragraph{Normalization for evaluation.}
Raw answerlines and prompts contain encoding noise and formatting conventions that can create spurious evaluation failures for both humans and models. We apply multi-stage cleaning that targets:
\textit{(1) Character normalization:} mapping diacritics and special symbols to ASCII equivalents and removing Unicode artifacts introduced by heterogeneous source encodings. 
\textit{(2) Formatting normalization:} stripping QuizBowl markup, removing prompt instructions and bracketed editorial artifacts, and rewriting alternative acceptable answers into a consistent ``A or B'' form.
\textit{(3) Context-aware cleanup:} for cases where rule-based edits are insufficient, we use GPT-4o-mini to rewrite answers into a standardized, human-readable form while preserving semantic content and listed alternatives. When cleanup is uncertain, we prefer conservative edits that preserve the original label over aggressive rewriting.

GPT-4o-mini was used strictly for \textbf{format normalization and controlled answer canonicalization}, not for modifying semantic content or introducing new answer keys. Specifically, its role was limited to:

\begin{enumerate}
    \item Removing encoding artifacts and markup (e.g., QuizBowl formatting such as underlines and bracketed instructions).
    \item Converting acceptance instructions into explicit canonical surface forms.
\end{enumerate}

For example, the original annotation:

\begin{quote}
\emph{``{``Pathétique'' Sonata} [or {Beethoven’s Piano Sonata No.~8} in C minor, {Op.~13} (accept any underlined part)]''}
\end{quote}

was normalized to:

\begin{quote}
\emph{``Pathetique Sonata or Beethoven's Piano Sonata No.~8 in C minor or Sonata No.~8 in C minor''}
\end{quote}

No new aliases were added beyond those explicitly licensed by the original annotation (``accept any underlined part''), and no semantic reinterpretation was performed. The model was used only to rewrite formatting artifacts into clean, evaluation-ready strings while preserving the original acceptance set.
\begin{table}[!tb]
\centering
\tiny
\begin{tabular}{p{2.2cm} p{0.8 cm} p{0.8cm} p{0.8cm} p{0.8cm}}
\toprule
Statistic & Pavements & Audio-Packets & Quizmasters & External Sources \\
\midrule
Questions (\%) & 6.94\% & 17.02\% & 42.70\% & 33.33\% \\
Avg. Audio Duration (s) & 63.42 & 65.25 & 41.81 & 13.75 \\
Avg. Ques. Length & 6.56 & 4.68 & 14.54 & 12.04 \\
\bottomrule
\end{tabular}
\caption{Distribution of questions in the \name{} Benchmark Dataset by source, with audio duration and question length statistics.}
\label{tab:audita_sources}
\vspace{-0.3cm}
\end{table}

\begin{table*}[!tb]
\centering
\tiny
\begin{tabular}{p{2.8cm} p{1.5cm} p{1.5cm} p{1.5cm} p{1.5cm} p{1.5cm} p{2cm}}\\
\toprule
Statistic & Who's Who? Name That Persona &Cultural Geography in Sound&Name The Music: Songs, Artists \& Composers&Pop Culture and Media&Elements of Musical Works&Environmental and Acoustic Sound Recognition\\
\midrule
\% Questions & 7.66\% & 4.56\% & 26.60\% & 19.13\% & 7.80\% & 34.24\% \\
Avg. Audio Duration (s) & 21.25 & 64.14 & 41.63 & 68.95 & 58.70 & 10.43 \\
Avg. Question Length (words) & 13.41 & 8.21 & 9.86 & 13.94 & 8.56 & 25.27 \\
\bottomrule
\end{tabular}
\caption{Breakdown of questions in the \name{} Benchmark Dataset by category, with average audio durations and question lengths.}
\label{tab:audita_categories}
\vspace{-0.3cm}
\end{table*}

\begin{table*}[t]
\centering
\tiny
\setlength{\tabcolsep}{6pt}
\begin{tabular}{p{0.12\linewidth} p{0.28\linewidth} p{0.28\linewidth} p{0.26\linewidth}}
\hline
\textbf{Family} & \textbf{Typical Construction} & \textbf{Common Shortcut and What \name{} Stresses} & \textbf{Typical Question Style \& Example} \\
\hline
Sound labeling & Closed label sets, short clips, event tags & Salient cue detection, weak long context. \name{} uses longer clips and open answer space. & \emph{Example:} Classify short environmental sounds (e.g., engine, applause). \\
Caption-derived QA & Questions derived from captions or metadata (e.g., OpenAQA~\cite{ltu}) & Lexical priors and caption artifacts. \name{} uses human-authored trivia not derived from captions. & Caption-consistent attribute query. \emph{Example:} Identify the sound source described by the caption. \\
Synthetic or templated & Generated scenes, fixed templates, constrained language & Template regularities and limited linguistic diversity. \name{} uses natural, non-templated questions. & Programmatic logic over fixed attributes. \emph{Example:} Count occurrences of a specified event. \\
Speech-centric QA & Spoken content dominates, transcript-like supervision & Can collapse to ASR plus text QA. \name{} includes speech but also music and non-speech audio. & Spoken content identification, often transcript-based questions. \\
\name{} & Human-written questions with real-world referents & Requires audio entity linking and multi-cue integration. This is the core target regime of \name{}. & Probing trivia grounded in real referents. \emph{Example:} Identify a film theme, speaker, or cultural artifact from audio cues. \\
\hline
\end{tabular}
\caption{Positioning of \name{} relative to common AQA benchmark designs, including typical dataset construction, common shortcuts that models exploit, and representative question styles and examples.}
\label{tab:merged_positioning_examples}
\end{table*}

\begin{table*}[h]
\centering
\tiny
\begin{tabular}{p{2cm} p{4cm}p{3.5cm} p{1.5cm} p{3cm}}
\toprule
\textbf{Dataset} & \textbf{Example Question} & \textbf{Answer (Gold)} &\textbf{IRT Difficulty / Discrimination} & \textbf{Issue / Notes} \\
\midrule

VGGSound QA~\cite{DBLP:conf/icassp/ChenXVZ20} &
\q{What type of animal is making the high-pitched and sharp sound described in the audio?} &\emph{The high-pitched and sharp sound is most likely a bark produced by a dog.}&
Low difficulty, low discrimination ($b=-2.1,\ a=0.35$) &
Simple audio classification task. Some clips have overlapping sounds causing ambiguity. Occasional metadata leakage possible, making some questions answerable without listening.\label{row1} \\

\midrule

ClothoAQA~\cite{DBLP:conf/icassp/DrossosLV20} &
\q{Is this outdoors?} &\emph{Yes}&
Moderate difficulty, low discrimination ($b=-0.6,\ a=0.42$) &
Binary classification style question, with limited complexity. Contextual metadata sometimes gives away the answer. Not all questions require detailed auditory reasoning. \\

\midrule

AudioCaps QA~\cite{DBLP:conf/naacl/KimKLK19} &
\q{Create a brief audio description, create labels, caption next.} &\emph{Labels: Vehicle; Tire squeal; Car; Race car, auto racing. Audio caption: Race car engines speed by, changing gears and screeching.} &
Varied difficulty, often low discrimination ($b\approx0.1,\ a=0.28$)&
Open-ended captioning questions. Subjective answers complicate evaluation and lack precise metrics. Do not constitute discrete QA. \\

\midrule
``No listening needed'' Questions (e.g. AudioCaps QA~\cite{DBLP:conf/naacl/KimKLK19}) & 
Examples: \q{What is the most likely reason for someone to strike a metal trailer with a wooden rod?} or \q{What is the significance of thunder in mythology?}&\emph{There could be a variety of reasons, such as trying to get someone's attention, testing the durability of the trailer, or making a musical sound.} and \emph{In many cultures, thunder represents the authority of gods and goddesses, and it is often associated with power, strength, and fertility.}&
Very low difficulty, near-zero discrimination ($b=-3.1,\ a\approx0$) and ($b=-3.4,\ a\approx0$)&
Answerable without listening, often appearing in scraped or poorly filtered datasets. Undermines auditory reasoning benchmarks. Not typical of curated datasets but important to highlight. \\
\bottomrule
\end{tabular}
\caption{Representative question examples from popular audio QA datasets, annotated with illustrative ranges of psychometric (IRT) difficulty and discrimination, and highlighting issues such as reliance on metadata, low reasoning complexity, and ``no listening needed'' questions. This underscores the need for carefully curated, human-authored datasets that robustly evaluate auditory reasoning.}
\label{tab:dataset_issues_real}
\end{table*}
\begin{table*}[h]
\centering
\tiny
\begin{tabular}{p{3cm} p{3cm} p{4cm} p{5cm}}
\toprule
\textbf{Issue} & \textbf{Dataset(s)} & \textbf{Example Question \& Answer}  & \textbf{Notes} \\
\midrule

Ambiguity & Clotho-AQA~\cite{DBLP:conf/icassp/DrossosLV20}, VGGSound QA~\cite{DBLP:conf/icassp/ChenXVZ20}, AudioCaps QA~\cite{DBLP:conf/naacl/KimKLK19}, FSD50K QA~\cite{DBLP:journals/taslp/FonsecaFPFS22} & 
Q: \q{What animal is making the sound?} \newline
A: \ans{bird} / \ans{dog} (multiple audible) \newline
Q: \q{Where is the sound coming from?} \newline
A: \ans{indoors} / \ans{outside} (disagreement among participants) \newline
Q: \q{What is the dominant sound in the clip?} \newline
A: \ans{siren} / \ans{car horn} (subjective) \newline
Q: \q{Is there a sound of laughter?} \newline
A: \ans{yes} / \ans{no} (faintness ambiguity) \newline
Q: \q{Is there a vehicle sound?} \newline
A: \ans{yes} / \ans{no} (ambiguous engine/horn sounds)  & 
Multiple overlapping sounds cause unclear targets; vague or underspecified question wording leads to inconsistent answers across annotators and models. \\

Weak Grounding / Shortcut & AudioCaps QA~\cite{DBLP:conf/naacl/KimKLK19}, AudioSet QA~\cite{DBLP:conf/icassp/GemmekeEFJLMPR17} & 
Q: \q{Is someone laughing?} \newline A: \ans{yes} \newline
Q: \q{Is there music playing?} \newline A: \ans{yes}  & 
Questions answerable from captions or metadata alone without listening; templated language encourages shortcut learning. \\

Underspecified Answer Keys & MUSIC-21 QA~\cite{DBLP:journals/tismir/ChristodoulouGLJ25}, Clotho-AQA~\cite{DBLP:conf/icassp/DrossosLV20} & 
Q: \q{What instrument is playing?} \newline A: \ans{piano} \newline
Q: \q{What animal can be heard?} \newline A: \ans{dog} / \ans{puppy} / \ans{hound}  & 
Multiple valid lexical variants treated as separate answers, increasing noise. \\

False Presuppositions & Speech Commands QA~\cite{DBLP:journals/corr/abs-1804-03209} & 
Q: \q{Is the command `stop' present?} \newline A: \ans{no} & 
Questions assume presence of commands that may not exist, confusing annotators and models. \\

Overlapping Audio & VGGSound QA~\cite{DBLP:conf/icassp/ChenXVZ20}, AudioCaps QA~\cite{DBLP:conf/naacl/KimKLK19} & 
Q: \q{What animal is making the sound?} \newline A: \ans{dog} &  
Overlapping sound sources cause ambiguity, complicating correct labeling and answering. \\

Synthetic Question Bias & AudioSet QA~\cite{DBLP:conf/icassp/GemmekeEFJLMPR17} & 
Q: \q{Is there music playing?} \newline A: \ans{yes} & 
Automated templated generation reduces linguistic diversity and causes models to exploit shortcuts. \\

\bottomrule
\end{tabular}
\caption{Summary of common issues in audio QA datasets, including improved, concrete ambiguous question examples from real datasets.}
\label{tab:dataset_issues_real_improved2}
\end{table*}
\section{OpenAQA Filtering and ClothoAQA Evaluation Setup}\label{app:dataset-positioning}
\paragraph{OpenAQA Filtering}
The 18.65\% removal follows the filtering script released by the original OpenAQA authors~\cite{DBLP:conf/iclr/0001LLKG24}. This script excludes GPT-generated responses that explicitly signal insufficient information, such as phrases like ``cannot be determined,'' ``not provided,'' ``unclear,'' and related hedging expressions. Importantly, this is not a subjective quality-based filter; rather, it removes instances where the generative pipeline itself identified the question as unanswerable. The remaining 81.35\% therefore consist of questions paired with committed, specific answers.

\paragraph{Positioning ClothoAQA Within Our Evaluation Framework} ClothoAQA occupies a distinct position in our evaluation framework. Our critique focuses on its structural limitations as a standalone benchmark, including ambiguous questions, overlapping audio sources, and underspecified answer keys. At the same time, ClothoAQA is crowd-sourced from human annotators rather than synthetically generated, placing it in a meaningfully different quality tier compared to caption-derived or templated datasets. Moreover, it was not included in OpenAQA, making it a genuinely independent data source. We include ClothoAQA to provide human performance baselines on a dataset already used by the community—not because we consider it free from the limitations discussed in Section~\ref{sec:motivation}.

Across the dataset, human evaluation collected $1517$ human guesses-individual answer attempts by participants- providing a reliable set of judgments to benchmark model performance. In the next section, we describe our evaluation framework, which leverages these human guesses to jointly model question properties and participant abilities.
\section{Answer Formats}\label{answer_format}

We evaluate two answer settings:
\paragraph{Free-Response Question Answering.}
In the free-response setting, models generate open-ended textual answers. Generated responses are evaluated using the PEDANTS~\cite{DBLP:conf/emnlp/LiMNLB24} framework, which determines semantic equivalence between model outputs and reference answers through structured normalization and equivalence rules, rather than relying on exact string matching.

\paragraph{Multiple-Choice Question Answering (MCQ).}
In the MCQ setting, models are given a fixed set of candidate answers and must select the correct option. We convert model outputs to a choice via either explicit option selection or scoring each candidate independently, depending on model capabilities. Accuracy is reported as the fraction of correctly answered questions.
\input{2025_arr_audio/sections/dataset_com}
\section{Evaluated Models: Architectures, Training Objectives, and Assumptions}
\label{app:model-details}
We evaluate a diverse set of state-of-the-art open checkpoint models runnable locally, covering mid-scale language backbones (approximately 6B to 17B parameters in the language component) and a variety of audio front ends. These include models trained primarily on audio–text alignment objectives, large multimodal foundation models with audio inputs, and audio-specialized models adapted for question answering. To address the varied nature of questions—some requiring speech understanding such as lyrics or quoted lines—we categorize models into three groups: (i) audio-only understanding, (ii) speech-aware models, and (iii) unified audio+speech models, reporting results separately for each subgroup. For all models, we use publicly released checkpoints and follow recommended inference settings when available, with no fine-tuning on our dataset.

We evaluate a diverse collection of state-of-the-art open-checkpoint models designed for audio and multimodal language understanding, spanning mid-scale language backbones ranging from approximately 6 billion to 17 billion parameters. These models differ widely in their training objectives, architectural designs, and audio processing strategies. Among them are models primarily trained for audio–text alignment, such as Qwen2.5-Omni~\cite{DBLP:journals/corr/abs-2503-20215} and AudioGPT~\cite{DBLP:conf/aaai/HuangLYSCYWHHLR24}, which focus on aligning acoustic features with language representations to handle speech and audio understanding tasks. In addition, we consider large multimodal foundation models that incorporate audio inputs alongside other modalities, including OpenOmni~\cite{DBLP:journals/corr/abs-2501-04561}, Audio-Flamingo~\cite{DBLP:conf/icml/KongGBPVC24}, and Phi-4-Multimodal~\cite{DBLP:journals/corr/abs-2503-01743}. These models leverage powerful language backbones integrated with audio encoders, enabling flexible multimodal reasoning across diverse input types. Specialized audio language models such as SALMONN-2 and SALMONN-2+~\cite{DBLP:journals/corr/abs-2506-15220} emphasize improved audio–language alignment and training methods to enhance reasoning over complex auditory inputs, while models like MU-LLaMA~\cite{DBLP:conf/icassp/LiuHSS24} and the original SALMONN~\cite{DBLP:conf/iclr/TangYSC000M024} also contribute to the landscape of advanced multimodal comprehension. Other models, including Qwen3-Omni~\cite{DBLP:journals/corr/abs-2509-17765}, LTU-AS~\cite{DBLP:conf/asru/GongLLKG23}, Baichuan-Omni-1.5~\cite{DBLP:journals/corr/abs-2501-15368}, and VITA-1.5~\cite{DBLP:journals/corr/abs-2501-01957}, further expand the diversity of architectures and training approaches assessed.

For each model, we report performance on both text-based and multiple-choice (MCQ) question formats from our dataset, including accuracy percentages and item response theory (IRT) estimated ability scores ($\theta$), which provide a latent measure of model proficiency relative to question difficulty. Models are ranked within each task format to facilitate comparative evaluation. All evaluations utilize publicly released checkpoints with recommended inference settings, without any task-specific fine-tuning, ensuring an unbiased benchmarking environment. This comprehensive evaluation enables us to analyze strengths and limitations across modalities, task types, and audio reasoning capabilities, thereby offering insights into current progress and challenges in the field of audio question answering.
\section{Human Evaluation and Participant Instructions}\label{human}

\textbf{Participant Summary}
\begin{itemize}
    \item \textbf{Data collected:} 1517 individual answer attempts from 87 participants recruited from trivia communities.
    
    \item \textbf{Participant background:} Participants were not pre-selected experts but were generally familiar with common trivia domains (e.g., music, film, and pop culture).
    
    \item \textbf{Dataset coverage:} The dataset includes questions across multiple categories, allowing participants to engage with questions aligned with their strengths. This ensures that maximum achievable accuracy is not constrained by any single individual’s expertise.
\end{itemize}

Overall, this evaluation provides a \textbf{robust estimate of human performance}, capturing both question difficulty and variability in participant knowledge.

We will provide detailed participant instructions in the Appendix and plan to release the collected human answers after blind review, supporting reproducibility and transparency.

Participants were also asked to select a category at the beginning of the task to ensure they answered questions within a domain they felt most comfortable with.

\textbf{Instructions Given to Participants}\label{ins}

\begin{enumerate}
    \item \textbf{Two chances to answer:}
    \begin{itemize}
        \item \textbf{First chance:} Type your answer freely in the text box provided.
        \item \textbf{Second chance:} Select one answer from the four multiple-choice options.
    \end{itemize}

    \item \textbf{Feedback page:} After completing both responses, participants see a feedback page showing:
    \begin{itemize}
        \item The correct answer
        \item Their text response
        \item Their multiple-choice (MCQ) selection
        \item Whether the text response and MCQ choice were correct or incorrect
    \end{itemize}

    \item \textbf{Performance tracking:}
    \begin{itemize}
        \item \textbf{Text Score:} Number of correct text responses
        \item \textbf{MCQ Score:} Number of correct multiple-choice responses
        \item Both scores are displayed at the end of the activity
    \end{itemize}

    \item \textbf{Audio timing:}
    \begin{itemize}
        \item Automatically tracked to record how long each audio clip is played
        \item Does \textbf{not} affect the final score
    \end{itemize}

    \item \textbf{Post-evaluation:}
    \begin{itemize}
        \item All text responses undergo human review to ensure fair and accurate scoring
    \end{itemize}
\end{enumerate}

\textbf{Tip:} Listen carefully, use headphones if possible, and try your best on both attempts.
\subsection{Participant List}\label{participant_list}

We thank all participants for their contributions to data collection and evaluation. All participants provided consent to be acknowledged (Table~\ref{tab:participants}).
\input{2025_arr_audio/sections/participant_list.tex}
\section{MCQ Analysis, Human Agreement, and Task Difficulty}
\label{app:mcq-analysis}
\paragraph{MCQ Output Class Distribution Analysis}

We analyzed the class distribution of model predictions across all four answer options to better understand convergence behavior and potential bias in the MCQ setting.

Aggregating across all models and items, the output distribution is shown in Table~\ref{tab:mcq_distribution}. Importantly, the model does not collapse onto a single dominant option, and predictions are distributed across all four classes. The distribution remains relatively balanced overall. This suggests that the below-chance MCQ accuracy (15.65\%) is not driven by trivial positional collapse or degenerate convergence behavior.

Instead, the pattern is more consistent with systematic selection of semantically plausible but incorrect distractors. In other words, the model is neither guessing uniformly at random nor defaulting to a single class; rather, it appears to rely on partial cues that lead to structured but incorrect choices (Table~\ref{tab:mcq_distribution}).

This analysis highlights that errors are distributed fairly evenly across distractors, further supporting that model failures arise from structured but incorrect semantic choices rather than simple positional bias.
\paragraph{Inter-Human Agreement}

Questions were answered by multiple participants, enabling aggregation of correct answers across individuals to approximate a reliable human topline.

Average agreement across participants varies by question type but demonstrates that questions are answerable given attention and domain knowledge. Inter-human agreement (top 20\% skilled participants, PEDANTS-normalized exact match) is {0.91} for open-ended questions ({1.0} in MCQ format), indicating very high convergence among the most consistent participants despite the open-ended format.

This supports that the questions are {well-posed and answerable}, while remaining challenging.

\paragraph{Human Topline Score and Task Difficulty}

The human topline score of {32\%} reflects the difficulty of open-ended audio reasoning, not infeasibility. All questions are collected from real-world trivia competitions, curated online tournaments, and expert-designed quizzes (Section~3).

These questions are explicitly intended to be answerable by humans, with each item verified and crafted to ensure solvability given attentive listening and relevant knowledge. For example, participants might be asked to identify a composer from a musical motif or an actor speaking in a clip.

Such tasks require careful {auditory perception} combined with {cultural or domain knowledge}. The 32\% score highlights the intrinsic challenge of precise auditory reasoning in open-ended text responses, rather than a lack of information (Table~\ref{task-dif}).

\paragraph{Explanation for Below-Chance MCQ Performance}

To better understand model behavior on multiple-choice questions, we analyzed prediction patterns, human agreement, and task difficulty (Appendix~\ref{app:mcq-analysis} for full tables and examples). Models achieve an average MCQ accuracy of 15.65\%, below the nominal 25\% chance level for four-option questions. This below-chance performance is not caused by dataset flaws or trivial positional collapse; instead, it arises from structured confusion among plausible distractors, reliance on subtle audio cues, and the inherently challenging nature of open-ended audio trivia. For instance, in the question \q{Name the title character of these movies} (correct answer: \emph{Batman}), the model selected \emph{Superman} among highly plausible alternatives, illustrating systematic but semantically reasonable errors.

Analysis of class distributions confirms that predictions are spread across all four options rather than collapsing to a single choice (see Table~\ref{tab:mcq_distribution} in Appendix~\ref{app:mcq-analysis}). Similarly, error distribution across distractors is relatively balanced, supporting the interpretation that models make structured mis-selections rather than random guesses.

Human validation further demonstrates that these questions are answerable: the top 20\% of participants achieve near-perfect agreement (1.0 in MCQ format), and the open-ended human topline reaches 32\%, highlighting the intrinsic difficulty of precise auditory reasoning rather than dataset ambiguity. Representative high- and low-difficulty examples are provided in Appendix~\ref{app:mcq-analysis}.

Taken together, these findings indicate that \name{} questions are challenging yet solvable, and that model failures primarily reflect the interplay of perceptual, semantic, and distractor-driven challenges rather than flaws in dataset construction.

The observed {MCQ accuracy of 15.65\%}, which is below the chance level of 25\% for 4-choice questions, arises from the combination of {extremely large answer spaces} in the open-ended audio trivia domain and the way distractors are designed.

\begin{enumerate}
    \item {Systematic confusion:} Many questions contain plausible distractors that are similar to the correct answer, leading models to select incorrect options more frequently than random guessing.
    
    \item {Question--audio alignment:} Some questions rely on subtle audio cues that the model cannot reliably perceive, further skewing predictions toward incorrect distractors.
    
    \item {High difficulty nature:} The dataset contains rare or niche knowledge (e.g., obscure songs, actors, or composers), which is challenging even for models with strong general language understanding, resulting in lower-than-chance accuracy in multiple-choice settings.
\end{enumerate}

Thus, the below-chance performance does {not indicate a quality issue with the dataset}, but rather reflects the combination of {carefully constructed distractors, high question difficulty, and reliance on perceptual audio understanding}, which models currently struggle to handle.

For example, consider the question: \emph{\q{Name the title character of these movies.}} The correct answer was \ans{Batman} (acceptable variants included \emph{Bruce Wayne} or \emph{Dark Knight}), with distractors \emph{Spider-Man}, \emph{Superman}, and \emph{Iron Man}. The model selected \emph{Superman}. All options are highly plausible superhero characters, and without confidently grounding the audio cues, the model gravitates toward semantically related but incorrect alternatives.

When such confusion occurs systematically across many items with strong distractors, accuracy can fall below the nominal 25\% chance level for four-option MCQs---not because of dataset flaws, but because models exhibit structured miscalibration rather than random guessing.

For reference, the {top 20\% of human participants achieve near-perfect agreement}, with MCQ responses scoring {1.0} (agreement) and an overall accuracy of {89.33\%}. This demonstrates that, while models struggle below chance, high-performing humans are able to reliably answer the same questions, confirming that the dataset is answerable and that the low model performance reflects task difficulty and distractor design rather than dataset flaws.

\begin{figure}[!tb]
    \centering
    \includegraphics[width=0.35\textwidth]{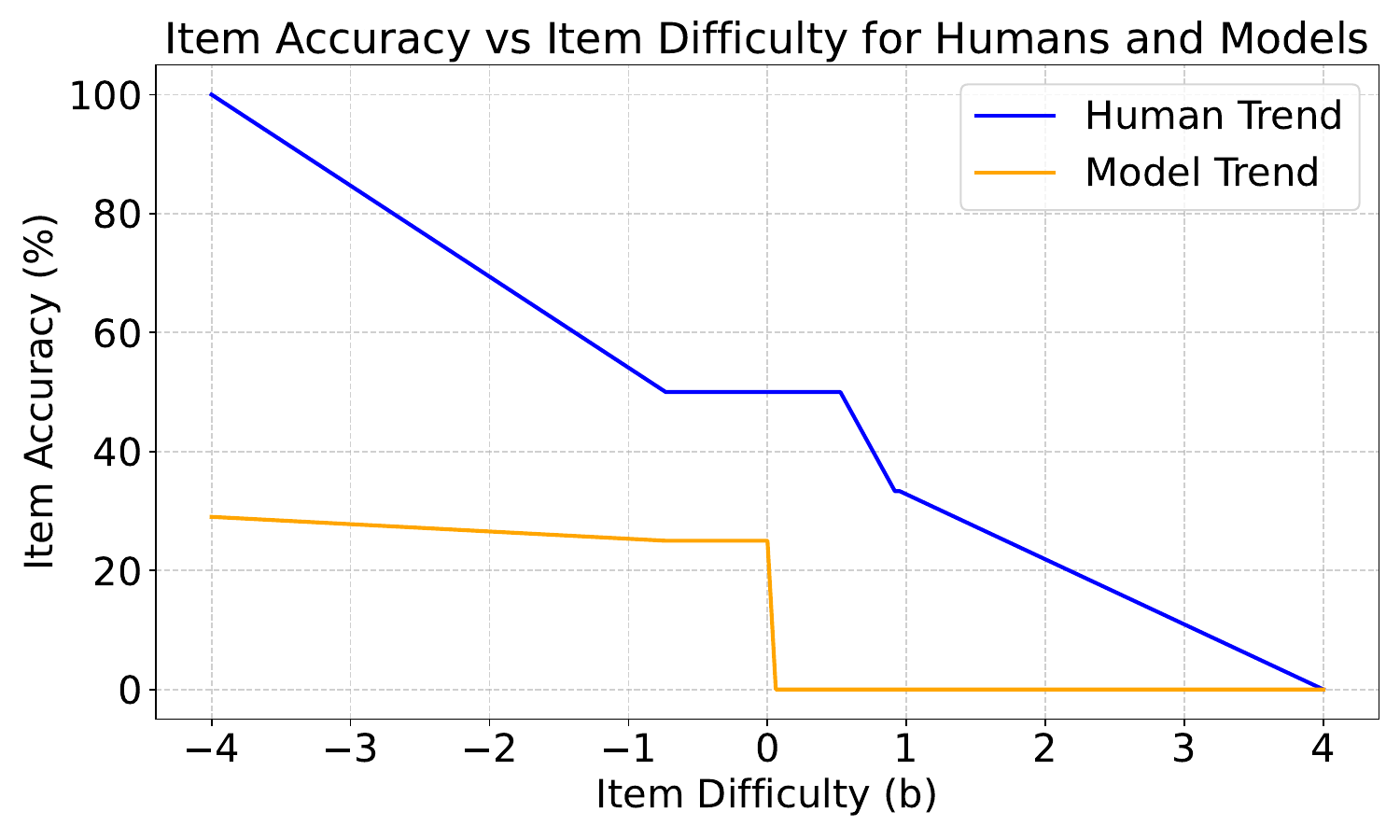}
    \caption{Item accuracy plotted against item difficulty (b) for humans (blue) and models (orange). Humans generally show higher accuracy than models across difficulties; however, some human accuracy values drop to zero due to sparsity of responses on certain items. The trend lines, highlight the widening performance gap between humans and models.}
    \label{fig:item}
\vspace{-0.4cm}
\end{figure}
\paragraph{Questions are Challenging but Gettable}

When a set of questions exhibits low accuracy, it is natural to suspect that the dataset may contain false presuppositions, ambiguities, or incorrect answer keys. One of the goals of the human validation process is to ensure that this is not the case for this dataset.

Item Response Theory (IRT) analysis provides additional verification. Most examples are answered correctly by high-skilled human participants, with only 26 examples never receiving a correct human response. Average agreement across participants varies by question type but demonstrates that questions are answerable given sufficient attention and domain knowledge.

Inter-human agreement (PEDANTS-normalized exact match) is {0.36} ({0.57} in MCQ format), indicating substantial convergence among participants despite the open-ended format. These results support that the questions are {well-posed and answerable}, while remaining challenging.
\section{Related Work}\label{app:related_work}
\paragraph{Audio question answering (\aqa{})}Audio Question Answering remains a nascent field with relatively few datasets, many of which impose design constraints that limit their ability to evaluate genuine auditory reasoning. Early efforts such as CLEAR~\cite{DBLP:conf/nips/LinSPR21} construct synthetic acoustic scenes by layering individual musical notes from the GoodSounds database. Questions are programmatically generated from logical templates and target specific attributes, for example, \q{How many times does the C note occur in this clip?} While this approach enables precise semantic control, it restricts linguistic diversity and limits reasoning complexity.

Similarly, DAQA~\cite{DBLP:journals/taslp/FayekJ20} composes variable-length audio clips from a closed vocabulary of 32 sound classes (e.g., ans{dog bark,} \ans{car horn}) and asks questions such as \q{Does the sound of a car horn occur more than twice in this clip?} Although DAQA allows limited temporal reasoning, its answer space is restricted to yes/no or counts, constraining the evaluation of richer auditory inference.

More naturalistic datasets attempt to move beyond synthetic audio. ClothoAQA~\cite{DBLP:conf/icassp/DrossosLV20} relies on crowd workers to write questions about environmental sound recordings originally collected for captioning. Questions such as \q{What animal makes the sound in this clip?} or \q{Is the sound recorded indoors or outdoors?} better resemble real-world queries. However, this process introduces strong linguistic priors: models trained on ClothoAQA perform competitively even when audio is removed, indicating reliance on textual cues rather than acoustic understanding. Music-AVQA~\cite{DBLP:conf/cvpr/LiWTXW022} exhibits similar limitations, using templates like \q{What instrument is playing?} or \q{Is the tempo fast or slow?} These formats further limit linguistic variability and encourage shortcut learning, echoing issues observed in textual and visual QA (Section~\ref{app:related_work}).

\paragraph{Audio Question Answering Benchmarks} Existing AQA benchmarks provide important testbeds for audio–language modeling but do not systematically expose failures in human-relevant auditory reasoning. CLEAR and ClothoAQA primarily reduce to controlled attribute queries or implicit classification tasks, making it difficult to distinguish true reasoning from surface-level pattern matching. While foundational, these benchmarks are limited in their ability to reveal nuanced model weaknesses in realistic, probing scenarios.

More recently, the DCASE 2025 Audio Question Answering challenge~\cite{DBLP:journals/corr/abs-2505-07365} introduced multi-domain QA subsets spanning bioacoustics, temporal soundscapes, and complex real-world audio. Although this effort broadens domain coverage, it remains centered on multiple-choice evaluation and lacks an adversarial, human-authored component designed to surface model brittleness.

Complementary evidence from \citet{DBLP:journals/corr/abs-2508-15407} shows that large audio–language models are highly sensitive to misleading or conflicting textual cues paired with audio, revealing robustness gaps in current evaluations. Together, these findings motivate benchmarks that deliberately incorporate adversarial examples and human-authored questions to stress robust audio–language reasoning.

\paragraph{Adversarial Evaluation in QA and Multimodal Tasks}Outside of audio QA specifically, adversarial evaluation has been successfully applied in other QA domains. In multimodal QA such as Visual Question Answering, human-in-the-loop adversarial data collection (e.g., Adversarial VQA) has been shown to produce questions that systematically expose model weaknesses by allowing annotators to target model failure modes through iterative feedback. Studies on adversarial QA in text also reveal that without adversarial examples, models can achieve high accuracy by exploiting dataset biases rather than robust reasoning. These findings underscore the value of adversarially collected questions for diagnosing model behavior, but comparable efforts are scarce for purely auditory content. 

To address these gaps, we construct a human-authored adversarial audio QA dataset grounded in real audio recordings and designed to be easy for humans but challenging for current models. Unlike prior AQA benchmarks, our dataset emphasizes semantic richness, adversarial focus, and natural realism, enabling evaluations that reveal model brittleness that structured or synthetic datasets fail to surface.
\textbf{Multimodal Question Answering} One of the first large-scale multimodal datasets for question answering was the VQA dataset~\cite{DBLP:conf/iccv/AntolALMBZP15}, which used image data to give models the context to answer a natural language question. The authors had crowd workers write questions about images from the COCO dataset and synthetically generated scenes to produce the data. Shortly after VQA was released in 2015, ~\citet{DBLP:conf/cvpr/JohnsonHMFZG17} of the CLEVR dataset created a VQA task that uses fully synthetic scenes to produce a comprehensive visual reasoning test~\cite{DBLP:conf/iccv/JohnsonHMHFZG17}. One of the key aspects of this dataset was that questions were generated using a functional program, which would inspire future Visual Datasets. 

One of the main issues within the VQA task was the presence of heavy priors within the data, where emergent statistical patterns would undermine the goal of reasoning over the image and the text. Many groups have been making efforts to improve upon this. The VQA v2 dataset balances the VQA dataset by introducing complements to each data point, where the new image is similar to the original but produces a different answer to the corresponding question~\cite{DBLP:conf/cvpr/GoyalKSBP17}. Later, efforts were made to control the distribution of answers by~\cite{DBLP:conf/cvpr/AgrawalBPK18} of the VQA-CP, who changed the splits of the VQA and VQA v2 datasets to alter the priors of the answer distribution. 
While remedies have been made to the original VQA dataset to fix its issues with heavy bias, other datasets have been introduced to overcome these pitfalls. One such dataset is GQA, which uses scene graphs based on images from COCO and Flickr to build questions automatically~\cite{DBLP:conf/cvpr/HudsonM19}. From the scene graph, questions and answers are built from a functional program similar to what was used in CLEVR, which further allowed the authors to smooth the answer distribution for various groups of questions. 

Outside of entirely image/text-based multimodal QA datasets, several examples of datasets explore different mediums and combine already popular ones. For example, the MultimodalQA creates multimodal questions by composing single modality questions about Wikipedia tables and the entities linked within them (such as images or other objects)~\cite{DBLP:conf/iclr/TalmorYCLWAIHB21}. To compensate for the algorithmic generation of questions, the authors use crowd workers to rephrase the question into a more natural alternative and have other workers verify the question's validity. Several video-based datasets have also been released following the pattern of looking at combinations of modalities. For example, ~\citet{DBLP:conf/mm/Yang0DCHJ022} used videos from the VGG-sound dataset and expert annotators to write questions about each video for AVQA dataset. Similarly, ~\citet{DBLP:conf/cvpr/LiWTXW022} released the Music-AVQA dataset by collecting YouTube videos of music performances and crowd workers produce questions that followed a predefined template. 

While many of these datasets utilize datasets from adjacent tasks of similar modality, only a subset of those available are web-curated. An advantage of our dataset in this field is that humans have already written the questions we collect outside of the context of our research, which minimizes much of the bias observed from directly using crowd workers to produce data for a benchmark. 

\textbf{Audio Question Answering} -- AQA datasets have been generally sparse over the past 6 years. Major contributions were either synthetically generated like DAQA~\cite{DBLP:journals/taslp/FayekJ20} and CLEAR~\cite{DBLP:conf/nips/LinSPR21} or reliant on crowd workers for annotations like ClothoAQA~\cite{DBLP:conf/icassp/DrossosLV20}. One of the earliest examples of a contemporary AQA dataset is the CLEAR dataset, which shares many similarities with the CLEVR dataset. In particular, the CLEAR datasets combines individual musical notes from the Good-Sounds database to generate an acoustic scene for a model to analyze. Like CLEVR, CLEAR's questions are constructed from templates represented as a logical tree with a functional program associated with them~\cite{DBLP:conf/nips/LinSPR21}. 

Using a similar methodology of combining smaller events, the DAQA dataset constructs audio-question pairs by stitching together several audio events. The main difference between the two is that DAQA uses events of variable length at various frequencies, so they can ask questions about how often a specific event occurs within a clip. While this dataset can help test a surface level of reasoning, the answer space is very small, with only 32 classes, with many answers being yes or no. Later, the Clotho AQA dataset was made, which takes advantage of the Clotho audio captioning dataset and uses crowd workers to produce new questions. Regarding crowd work, the resulting dataset may often contain heavy priors as quality control has been difficult. There have been several attempts to treat this issue in the VQA space~\cite{Anderson2017BottomUpAT}; however, this hasn't been extended to the world of audio yet. Notably, this issue also skews results from this dataset, as many of the experiments conducted simply answering the question yield higher performance than using the audio and the question.

\textbf{Adversarial Dataset Creation} -- As models grow increasingly complex, it becomes significantly more difficult to understand precisely why a model makes particular decisions during its inferences. A consequence of this trend is that understanding the weaknesses of a model turns into a hard task, especially when it comes to black-box models. The goal of adversarial dataset generation is to explore how robust models are to noisy data and to look at ways models underperform compared to humans. Thus, an important quality of adversarial questions is that they are hard for computers but relatively easy for humans; if it is very difficult for both groups, then it simply shows the dataset may be too tough. For VQA, a group of researchers made Adversarial VQA (AVQA), which uses a human-in-the-loop approach to generate questions that challenge models. An important aspect of their question creation interface is that a SOTA VQA model is present to provide answer feedback, so people can tweak questions until the model is tricked~\cite{DBLP:conf/nips/ShengSGMTGPK21}. After the questions have been written, the model is retrained and given to the workers to write new, more difficult questions. In an interesting case of coincidence, within the same year,~\citet{DBLP:conf/iccv/LiLG021} produced a dataset called AdVQA using a similar technique. Although their annotation processes differ, AdVQA doesn't use model retraining. An interesting example of intentionally designing a challenging dataset is TriviaQA, a reading comprehension dataset that grounds its queries in trivia questions~\cite{DBLP:conf/acl/JoshiCWZ17}. Since Trivia questions are designed to test the ability of humans, they are also a great way to benchmark models, as they are written to find out the best of a group of people. 
\paragraph{Item Response Theory in NLP evaluation}
Item Response Theory (IRT), originally developed in psychometrics, has recently gained traction in NLP as a framework for evaluating both datasets and models. Early work by \citet{lalor2016building} introduced IRT to NLP by constructing evaluation sets for Recognizing Textual Entailment (RTE), demonstrating that standard metrics such as accuracy assume all items are equally informative, whereas IRT models item-specific properties such as difficulty and discrimination. This allows evaluation to account for heterogeneity in question quality and provides a more nuanced measure of system performance relative to human populations.

Subsequent work has expanded IRT beyond dataset construction to broader evaluation and analysis. \citet{lalor2020learning} show that IRT parameters can be estimated using model-generated responses (``artificial crowds''), enabling scalable application without extensive human annotation. More recent efforts apply IRT to diagnose biases and failure modes in NLP systems, using item parameters to quantify how model performance varies across demographic or linguistic factors \cite{xu2025fairness}.

Recent studies further highlight the role of IRT in rethinking benchmark evaluation for large language models. For example, \citet{zhou2026lost} argue that traditional leaderboards often fail to distinguish top-performing models due to poorly calibrated or low-discrimination items, and propose IRT-based frameworks to improve benchmark reliability and separability.
In parallel, tutorials and surveys (e.g., \citet{lalor2024item}) emphasize the growing adoption of IRT as a general tool for analyzing dataset quality, estimating model ability ($\theta$), and identifying informative versus uninformative examples. 

Overall, this line of work shows that IRT provides a principled alternative to aggregate metrics by jointly modeling item characteristics and system ability, enabling more reliable comparison, diagnostic analysis, and benchmark design.
\section{Result}
\subsection{Model Scale and Validity of Conclusions}
\label{app:scale_validity}
Many audio-language models couple a pretrained audio encoder to a pretrained text LLM by projecting audio features into the LLM token space, so the language backbone is a key determinant of downstream instruction-following and knowledge-heavy QA performance~\cite{ltu}. Evidence from recent scaling analyses of speech-text and audio-centric language models also supports the general trend that larger backbones and stronger decoders improve aggregate performance across understanding and reasoning tasks, although gains vary by task and domain~\cite{scaling_speech_text}.

At the same time, scale alone does not necessarily resolve failures caused by weak audio grounding and over-reliance on textual priors, which can manifest as confident but incorrect answers when audio evidence is insufficient or ignored~\cite{textbias}. This is consistent with MMAU, which reports that even strong proprietary systems remain far below human performance on its human-evaluated test-mini split and are only moderately separated from strong open models on the benchmark evaluation~\cite{sakshi2024mmaumassivemultitaskaudio}. For example, MMAU reports human performance of about 82.23 on test-mini, while Gemini Pro v1.5 achieves 52.97 on the test split and strong open models such as Qwen2-Audio-Instruct are comparable in several settings~\cite{sakshi2024mmaumassivemultitaskaudio}. These gains are meaningful, but they are not of a magnitude that would plausibly convert near-chance behavior on a difficult, human-authored benchmark into reliable audio reasoning.

Accordingly, while our evaluation focuses on open models in the mid-scale regime, our qualitative conclusions about failure modes and dataset difficulty are unlikely to be artifacts of model scale alone. Exhaustively evaluating cloud-scale proprietary models across all 9{,}690 questions would also impose substantial cost, which limits full scaling sweeps in this study.

\begin{table}[!tb]
\centering
\tiny
\begin{tabular}{p{1.75cm} p{3.75cm} p{1cm}}
\toprule
{Difficulty} & {Question} & {Answer} \\
\midrule
High (Low feasible) & What country is this national anthem from? & Bolivia \\
Low (High feasible) & What is the language spoken in this clip? & Chinese \\
\bottomrule
\end{tabular}
\caption{Example questions illustrating task difficulty.}
\label{task-dif}
\end{table}
\begin{table}[t]
\centering
\tiny
\begin{tabular}{p{2cm} p{2.5cm} p{2.25cm}}
\hline
{Option} & {Prediction Distribution (\%)} & {Error Distribution (\%)} \\
\hline
Option 1 & 25.45 & 24.90 \\
Option 2 & 29.16 & 28.92 \\
Option 3 & 24.65 & 24.52 \\
Option 4 & 20.74 & 21.66 \\
\hline
\end{tabular}
\caption{Model output distribution across MCQ answer options compared with distractor-level error distribution aggregated across all models. Errors are distributed fairly evenly across options, suggesting that below-chance performance is not caused by positional collapse.}
\label{tab:mcq_distribution}
\end{table}
\begin{table*}[t]
\centering
\small
\setlength{\tabcolsep}{8pt}
\begin{tabular}{rp{9cm}p{5cm}}
\toprule
\multicolumn{3}{c}{\textbf{Top 10 Best Discriminator Questions (IRT)}} \\
\midrule
\# & \textbf{Question} & \textbf{Answer}  \\
\midrule
1 & Name the title character of these movies. & \textit{Sherlock Holmes}  \\
2 & What is the name of the person who is speaking in this clip? & \textit{Harry Styles}  \\
3 & Name the character that inspired this music.&	\textit{Batman} \\
4&Name the lead artist.	&\textit{Halsey}\\
5 & Give the common word found in the names of these lead artists.&\textit{	A\$AP}  \\
6 & Name the artist.&	\textit{beabadoobee}  \\
7 & What country is this national anthem from?	&\textit{Bolivia} \\
8&What TV show is this clip from?	&\textit{A Team}\\
9&What TV show is this clip from?	&\textit{Happy days}\\
10&Name the city where these movies are wholly or mostly set	&\textit{Paris}\\
\bottomrule
\end{tabular}

\vspace{1em}

\begin{tabular}{rp{9cm}p{5cm}}
\toprule
\multicolumn{3}{c}{\textbf{Top 10 Worst Discriminator Questions (IRT)}} \\
\midrule
\# & \textbf{Question} & \textbf{Answer}  \\
\midrule
1 & What is the language spoken in this clip?&	\textit{Chinese} \\
2 & What is the language spoken in this clip?&	\textit{German} \\
3&What is the next line of lyrics that occurs after the song in the clip ends?	&\textit{Two and Two Were Four}\\
4&What is the next line of lyrics that occurs after the song in the clip ends?	&\textit{On the Pages in Between}\\
5&What country is this national anthem from?	&\textit{Iceland}\\
6&Name the character.	&\textit{Siegfried}\\
7&Name the mythical figure who is singing in these excerpts.	&\textit{Hades}\\ 
8&Name the male lead of these movies	&\textit{Charlie Chaplin}\\
9&What type of role do these characters have in common?	&\textit{Trouser role or Pants role or Breeches role or Male roles played by women}\\
10&What is the name of the person who is speaking in this clip?	&\textit{Bruce Forsyth}\\
\bottomrule
\end{tabular}
\caption{Examples of the top 10 best and worst discriminator questions by Item Response Theory (IRT), including their answers. High discrimination indicates questions that effectively differentiate between high- and low-ability respondents, while low discrimination indicates poor differentiation power.}
\label{tab:irt_discrimination_examples}
\end{table*}

\begin{table*}[t]
\centering
\small
\setlength{\tabcolsep}{3pt}
\renewcommand{\arraystretch}{1.12}
\begin{tabular}{p{0.2\linewidth}p{0.12\linewidth}p{0.22\linewidth}p{0.12\linewidth}p{0.26\linewidth}}
\toprule
Model & LLM Core & Inputs & Outputs & Rationale \\
\midrule
\multicolumn{5}{l}{\textit{Omnimodal models (6)}} \\
\midrule
Qwen2.5-Omni & 7B & text, audio, image, video & text, speech & Thinker--Talker architecture with explicit reasoning--speech decoupling and unified multimodal support. \\
Qwen3-Omni & 30B-A3B & text, audio, image, video & text, speech & MoE-based omni model (30B total, $\sim$3B active) optimized for scalable multimodal reasoning and streaming speech. \\
OpenOmni & 7B & text, audio, image & text, speech & Language-pivot alignment with progressive modality training and preference-based speech tuning. \\
VITA-1.5 & 7B & text, audio, image, video & text, speech & End-to-end omni model using a three-stage training pipeline without cascaded ASR/TTS. \\
Mini-Omni2 & 0.5B & text, audio, image & text, speech & Lightweight end-to-end omni assistant with parallel text--audio decoding and command-based duplex interruption. \\
Baichuan-Omni-1.5 & 7B & text, audio, image, video & text, speech & Unified decoder with explicit audio token modeling and staged multimodal training, optimized for Chinese--English bilingual use. \\
\midrule
\multicolumn{5}{l}{\textit{Audio-language models (4)}} \\
\midrule
Audio-Flamingo & 1.3B & text, audio & text & Flamingo-style gated cross-attention, sliding-window audio features, ICL/RAG and multi-turn dialogue. \\
Qwen2-Audio & 7B & text, audio & text & Whisper-large-v3--initialized audio encoder into Qwen-7B, unified prompting, SFT+DPO alignment. \\
LTU-AS & 7B & audio, text & text & Frozen Whisper perception + TLTR time/layer aggregation, continuous audio tokens + transcript, LLaMA-7B w/ LoRA. \\
MU-LLaMA & 7B & text, audio & text & Frozen MERT music encoder + adapter injection into LLaMA-2 7B, MusicQA supervision for QA and captioning. \\
\bottomrule
\end{tabular}
\caption{Evaluated models organized by capability grouping (Part 1 of 2). All models evaluated in audio-question to text-answer setting.}
\label{tab:model-details-1}
\end{table*}


\begin{table*}[t]
\centering
\small
\setlength{\tabcolsep}{3pt}
\renewcommand{\arraystretch}{1.12}
\begin{tabular}{p{0.2\linewidth}p{0.12\linewidth}p{0.22\linewidth}p{0.14\linewidth}p{0.26\linewidth}}
\toprule
Model & LLM Core & Inputs & Outputs & Rationale \\
\midrule
\multicolumn{5}{l}{\textit{Speech-capable models (6)}} \\
\midrule
Phi-4-Multimodal & 3.8B & text, audio, image, video & text & Mixture-of-LoRAs with frozen language backbone, modality-specific adapters, 128K context. \\
SpeechGPT & 13B & text, speech & text, speech & Discrete speech units expanded into LLaMA vocabulary, three-stage training with Chain-of-Modality instruction tuning. \\
AudioGPT & modular & text, speech, audio, music, image & text, audio, video & Modular orchestration system using ChatGPT to coordinate 16+ foundation models via ASR/TTS interface. \\
SALMONN & 13B & text, speech, audio, music & text & Dual encoder (Whisper + BEATs), window-level Q-Former, activation tuning for emergent abilities. \\
video-SALMONN 2 & 7B & text, audio, video & text & Frozen backbone with audio branch, MrDPO for caption optimization, atomic event-based quality metrics. \\
video-SALMONN 2+ & 7B & text, audio, video & text & Caption-enhanced training via MrDPO-generated data, SOTA on Video-MME/WorldSense/AVUT benchmarks. \\
\bottomrule
\end{tabular}
\caption{Evaluated models organized by capability grouping (Part 2 of 2). All models evaluated in audio-question to text-answer setting.}
\label{tab:model-details-2}
\end{table*}

\begin{table*}[t]
\centering
\tiny
\begin{tabular}{p{4cm} p{7cm} p{5cm}}
\toprule
\textbf{Dataset / Paper} & \textbf{Example or Description} & \textbf{Why Models Excel}  \\
\midrule
VGGSound~\cite{DBLP:conf/icassp/ChenXVZ20} & Models detect synthetic or repeated alarm/beep sounds perfectly. These sounds have highly structured, repetitive waveforms easy for pattern matching. & Models trained on millions of audio clips memorize these patterns and detect them with near-perfect accuracy, often better than non-expert humans.  \\

Speech Command Recognition (benchmark dataset)~\cite{DBLP:journals/corr/abs-1804-03209} & Recognizing isolated spoken command keywords like ``stop,'' ``go,'' or ``yes'' — models achieve $>$99\% accuracy, often exceeding average human recognition. & Limited vocabulary and clean synthetic data make these tasks trivial for models. \\

AudioSet Tagging~\cite{DBLP:conf/icassp/GemmekeEFJLMPR17} & Models detect environmental sounds like sirens, horns, or machine noises with very high precision, sometimes outperforming humans in noisy clips. & Large training data and strong feature extraction enable models to spot subtle acoustic cues missed by humans. \\
\bottomrule
\end{tabular}
\caption{Examples of tasks where models demonstrate superhuman or near-superhuman performance in audio question answering or classification.}
\label{tab:superhuman-model-performance}
\end{table*}
\section{Understanding Model Failures Across Modalities}
\subsection{Factual Knowledge vs.\ Audio Understanding}\label{app:factual-audio}

We agree that certain questions (e.g., \q{Name the artist who recorded this song} or \q{What film is this theme from?}) require both perceptual recognition and world knowledge. To directly address this concern, we conducted additional analyses leveraging two model settings already available in our experiments: Question + Transcript (Whisper output) and Question + Raw Audio.

This comparison allows us to disentangle whether performance differences stem from: Lack of perceptual audio understanding, Imperfect transcription, Insufficient factual/world knowledge. Specifically: If a model succeeds with the transcript but fails with the audio, this indicates perceptual audio processing limitations. If it fails in both settings, the bottleneck is more likely knowledge-based. If it succeeds with audio but not transcript, this suggests information present in non-verbal acoustic cues (e.g., melody, timbre, instrumentation) that transcripts cannot capture.

Importantly, accuracy is substantially higher in the {raw-audio condition (e.g., for GPT-4o 14.87\%)} than in the {transcript condition (e.g., for GPT-4o 7.26\%)}. If performance were primarily driven by factual recall from textual cues, we would expect transcript-based performance to match or exceed audio-based performance. Instead, the lower transcript accuracy suggests that automatic transcriptions omit or distort critical acoustic information necessary for correct identification.

Our results show that performance gaps persist even when transcripts are provided, indicating that failures cannot be attributed solely to missing encyclopedic knowledge. Moreover, the gap between transcript and raw-audio conditions highlights genuine audio-understanding limitations, not merely factual recall deficiencies (Table~\ref{tab13}).

\begin{table*}[t]
\centering
\tiny
\begin{tabular}{p{2.5cm} p{3cm} p{1cm} p{1cm} p{1cm} p{5cm}}
\toprule
{Case} & {Question} & {Answer} & {Transcript Result} & {Audio Result} & {Explanation} \\
\midrule

Transcript succeeds, audio fails (perceptual limitation) 
& Name the songwriter 
& Joni Mitchell (\textit{River}) 
& $\checkmark$ 
& $\times$ 
& Transcript contains lines from the song, allowing the correct answer. Raw audio alone does not provide enough cues for the model to identify the song. \\

Audio succeeds, transcript fails (acoustic cue reliance) 
& Name the composer 
& Carl Orff 
& $\times$ 
& $\checkmark$ 
& Distinctive choruses and instrumental timbres in the audio enable the model to answer correctly; the transcript does not contain useful information. \\

Fails in both (knowledge limitation) 
& You are listening to the voice of a fictional character; what is this character's name (do not give the name of the actor/actress)? 
& Frank Spencer 
& $\times$ 
& $\times$ 
& Requires encyclopedic knowledge of TV characters; neither transcript nor audio alone is sufficient. \\

\bottomrule
\end{tabular}
\caption{Examples illustrating how transcript and raw-audio conditions reveal different failure modes: perceptual limitations, acoustic-cue reliance, and knowledge-based failures.}
\label{tab13}
\end{table*}

Performance gaps persist even when transcripts are provided, indicating that failures cannot be attributed solely to missing factual or world knowledge. The gap between transcript and raw-audio conditions highlights genuine audio-understanding limitations, not merely factual recall deficiencies. These examples illustrate how the two modalities (transcript vs.\ audio) contribute differently, enabling classification of failures as perceptual, knowledge-based, or audio-cue-driven.
\subsection{Failure cases analysis}\label{app:failure-analysis}
We conducted a systematic error analysis over all model errors and categorized them using the same diagnostic logic described in the rebuttal (audio succeeds / transcript fails; transcript succeeds / audio fails; both fail). The breakdown over error cases is as follows: {Knowledge-based errors (both transcript and audio fail):} 78.23\%, {Perceptual errors (transcript succeeds, audio fails):} 8.82\%, {Audio-cue errors (audio succeeds, transcript fails):} 12.95\%. Several clarifications are important for interpreting these numbers.

First, approximately 21.77\% of errors (8.82\% + 12.95\%) are directly attributable to audio-related limitations. These cases are not reducible to missing encyclopedic knowledge. Second, the predominance of knowledge-based failures is consistent with the overall low accuracies (e.g.,  14.87\% audio; 7.26\% transcript for GPT-4o). Many questions require identifying specific composers, performers, fictional characters, or cultural artifacts; even perfect perception would not suffice without the relevant world knowledge.

However, a substantial subset of transcript failures arises because the automatic transcription is {empty, fragmentary, or effectively gibberish} when the audio contains no linguistic content (e.g., instrumental music, sound effects, or non-verbal acoustic signatures). In these cases, the transcript condition is structurally disadvantaged: failure cannot be attributed purely to missing world knowledge, since the relevant signal exists only in the acoustic domain.

Representative examples are shown in Table~\ref{tab14}.

\begin{table*}[h]
\centering
\tiny
\begin{tabular}{p{2cm} p{1.5cm} p{2cm} p{8.5cm}}
\toprule
{Question} & {Gold Answer} & {Transcript} & {Model Answer} \\
\midrule

Name the work. &
Jurassic Park &
Girl ouch Look at me Thanks for watching &
``The transcription provided does not contain enough information to identify a specific work. It appears to be a mix of different languages and phrases, but without additional context or recognizable content, it's not possible to determine the name of a work based solely on this transcription.'' \\

What country is this national anthem from? &
Armenia &
You &
``I'm sorry, but the transcription provided does not contain enough information to determine which country's national anthem it is. Could you provide more details or context from the transcription?'' \\

Name the opera. &
Madama Butterfly &
(empty) &
(no response) \\

\bottomrule
\end{tabular}
\caption{These examples illustrate that transcript-only inputs often fail when the audio contains non-verbal or partially verbal content. Even when the model has access to textual output, crucial identifying cues such as melody, instrumentation, or anthem structure are lost in the transcript. Therefore, audio is essential for correct resolution in \name{}, and transcript degradation alone can explain a substantial portion of failures.}
\label{tab14}
\end{table*}

In each case, the transcript contains either no usable information or severely degraded content. The identifying signal (melody, orchestration, anthem structure) exists only in the acoustic domain. Thus, these failures reflect structural information loss in the transcript condition rather than purely missing world knowledge.

Overall, while knowledge-based failures account for 78.23\% of errors under our categorization scheme, this figure should not be interpreted as indicating that these failures are \textit{purely} knowledge-driven. The categories are analytically defined for diagnostic clarity, but in practice, many items involve {overlapping bottlenecks}.

In particular, a substantial subset of cases classified as ``knowledge errors'' occur in settings where the transcript is empty, fragmentary, or unintelligible due to the absence of linguistic content (e.g., instrumental passages, orchestral themes, non-verbal acoustic signals). In such cases, the transcript condition contains little or no recoverable evidence. While successful identification still ultimately requires world knowledge (e.g., recognizing a specific anthem or opera), the failure cannot be attributed to missing knowledge alone—the relevant perceptual signal is either unavailable or degraded in the transcript representation.

Thus, the 78.23\% figure reflects the \textit{final failure mode} (i.e., the model did not retrieve the correct entity), but many of these items are jointly constrained by:

\begin{itemize}
\item Knowledge requirements (e.g., knowing the composer, work, anthem, or character), and
\item Audio-specific limitations or structural signal loss (e.g., non-linguistic music that transcripts cannot encode).
\end{itemize}

Importantly, 21.77\% of errors are explicitly modality-driven (8.82\% perceptual; 12.95\% audio-cue), and an additional portion of the ``knowledge'' category includes items where transcript degradation meaningfully contributes to failure. Therefore, although knowledge is the largest single labeled category, the empirical picture is not one of purely encyclopedic deficiency. Rather, failures frequently arise from the interaction between knowledge demands and modality-specific constraints.

%% file: 2025_arr_audio/sections/dataset_com.tex
\begin{table}[t]
\centering
\tiny
\begin{tabular}{l l c c c}
\toprule
\multicolumn{5}{c}{\textbf{Text Questions}} \\
\midrule
Dataset & Modality & Acc (\%) & Mean $\theta$ & SD $\sigma$ \\
\midrule
Pavements & Human & 35.28 & 0.09 & 1.74 \\
Pavements & Model & 4.26 & -2.81 & 0.60 \\
\midrule
Audio-Packets & Human & 34.24 & 0.08 & 0.61 \\
Audio-Packets & Model & 8.89 & -2.03 & 0.67 \\
\midrule
Quizmasters & Human & 21.34 & 0.03 & 0.91 \\
Quizmasters & Model & 1.58 & -3.88 & 0.62 \\
\midrule
External Datasets & Human & 25.79 & 0.05 & 0.51 \\
External Datasets & Model & 13.49 & -1.45 & 0.51 \\
\midrule
\multicolumn{5}{c}{\textbf{MCQ Questions}} \\
\midrule
Dataset & Modality & Acc (\%) & Mean $\theta$ & SD $\sigma$ \\
\midrule
Pavements & Human & 53.90 & 0.11 & 1.75 \\
Pavements & Model & 5.89 & -2.33 & 0.61 \\
\midrule
Audio-Packets & Human & 61.21 & 0.13 & 0.62 \\
Audio-Packets & Model & 11.76 & -1.83 & 0.65 \\
\midrule
Quizmasters & Human & 50.31 & 0.10 & 0.93 \\
Quizmasters & Model & 3.04 & -2.70 & 0.62 \\
\midrule
External Datasets & Human & 74.30 & 0.15 & 0.49 \\
External Datasets & Model & 20.59 & -1.11 & 0.50 \\
\bottomrule
\end{tabular}
\caption{Comparison of accuracy and mean ability ($\theta$) across datasets, modalities (human vs. model), and question types (Text vs. MCQ) shows humans consistently outperform models. Pavements and Audio-Packets yield moderate human accuracy ($\approx$30–60\%) with positive abilities, while models score lower with negative abilities. Quizmasters is more challenging, especially for models. External datasets show the highest human accuracy, notably on MCQs. These results reveal clear human-model gaps and varying task difficulty by dataset and question type.}
\label{tab:text_mcq_split}
\vspace{-0.3cm}
\end{table}

%% file: 2025_arr_audio/sections/participant_list.tex
\begin{table*}[h]
\tiny
\begin{tabular}{lllllll}
\toprule
\multicolumn{7}{c}{\textbf{Participant Name}} \\
\midrule
Daniel Kim&Sara DelVillano &Raymond Kimball&Drew Scheeler&Nathan Zhao&Mohammed Afaan Mohammed Arif Ansari&Hemanth Nandakumar\\
Stefany Meyer & Jason Christopher&Forrest Weintraub&Nishant Balepur &Eve Nuria Fleisig\\
\bottomrule
\end{tabular}
\caption{List of participants who contributed to this study and agreed to be acknowledged by name.}
\label{tab:participants}
\end{table*}